%% file: s2vgd.tex
\documentclass[twoside]{article}
\usepackage{amsmath,amsthm,verbatim,amssymb,amsfonts,amscd, graphicx}
\usepackage{graphics, wrapfig}
\usepackage{wrapfig}
\usepackage[T1]{fontenc}

\usepackage{aecompl}
\newcommand{\av}{{\boldsymbol a}}

\newcommand{\dv}{{\boldsymbol d}}

\newcommand{\sv}{{\boldsymbol s}}

\newcommand{\vv}{{\boldsymbol v}}

\newcommand{\xv}{{\boldsymbol x}}
\newcommand{\yv}{{\boldsymbol y}}

\usepackage[bottom]{footmisc}

\newcommand{\Cmat}{{\bf C}}

\newcommand{\Hmat}{{\bf H}}
\newcommand{\Imat}{{\bf I}}

\newcommand{\Mmat}{{\bf M}}

\newcommand{\Pmat}{{\bf P}}
\newcommand{\Qmat}[0]{{{\bf Q}}}

\newcommand{\Smat}[0]{{{\bf S}}}

\newcommand{\Umat}[0]{{{\bf U}}}
\newcommand{\Vmat}[0]{{{\bf V}}}
\newcommand{\Wmat}[0]{{{\bf W}}}

\newcommand{\Ymat}{{\bf Y}}

\newcommand{\Lambdamat}{{\bf \Lambda}}

\newcommand{\thetav}{{\boldsymbol \theta}}

\usepackage[accepted]{aistats2018}
\usepackage{multirow}
\usepackage{tabularx}
\usepackage{ctable}
\usepackage{natbib}
\usepackage{indentfirst}
\usepackage{adjustbox}
\usepackage{bm}

\newcommand{\ie}[0]{\emph{i.e., }}

\newcommand{\Acal}{\mathcal{A}}

\newcommand{\Scal}{\mathcal{S}}

\newcommand{\E}{\mathbb{E}}

\def\vec{\textsf{vec}} 
\def\ReLU{\textsf{ReLU}} 
\def\KL{\textsf{KL}} 
\def\InvGamma{\textsf{Inv-Gamma}} 
\def\Softmax{\textsf{Softmax}} 
\def\Categorical{\textsf{Categorical}} 

\newtheorem{theorem}{Theorem}
\newtheorem{definition}[theorem]{Definition}
\newtheorem{lemma}[theorem]{Lemma}
\newtheorem{proposition}[theorem]{Proposition}

\theoremstyle{remark}

\newcommand{\RN}[1]{%
	\textup{\lowercase\expandafter{\it \romannumeral#1}}%
}

\usepackage{algorithm}
\usepackage{algpseudocode}
\usepackage{natbib}
\usepackage{hyperref}

\begin{document}

%

%

\twocolumn[

\aistatstitle{Learning Structural Weight Uncertainty for Sequential Decision-Making}

\aistatsauthor{  Ruiyi Zhang$^{1}$~~ Chunyuan Li$^{1}$~~  Changyou Chen$^{2}$~~ Lawrence Carin$^{1}$}

\aistatsaddress{ $^{1}$Duke University~~ $^{2}$University at Buffalo \\
  \texttt{ryzhang@cs.duke.edu,~cl319@duke.edu,~cchangyou@gmail.com,~lcarin@duke.edu}  }
]
\begin{abstract}
Learning probability distributions on the weights of neural networks (NNs) has recently proven beneficial in many applications. Bayesian methods, such as Stein variational gradient descent (SVGD), offer an elegant framework to reason about NN model uncertainty. However, by assuming independent Gaussian priors for the individual NN weights (as often applied), SVGD does not impose prior knowledge that there is often structural information (dependence) among weights. We propose efficient posterior learning of structural weight uncertainty, within an SVGD framework, by employing matrix variate Gaussian priors on NN parameters. We further investigate the learned structural uncertainty in sequential decision-making problems, including contextual bandits and reinforcement learning. Experiments on several synthetic and real datasets indicate the superiority of our model, compared with state-of-the-art methods.
\end{abstract}

\input{subtex/intro.tex}

\vspace{-3mm}
\input{subtex/preliminary.tex}
\input{subtex/methods.tex}

\input{subtex/experiments.tex}
\input{subtex/conclusion.tex}
\nocite{pu2017symmetric, zhang2016towards,zhang2017stochastic}
\paragraph{Acknowledgements}
We acknowledge Qiang Liu and Yang Liu for making their code public. We thank Chenyang Tao for proofreading. This research was supported in part by ARO, DARPA, DOE, NGA, ONR and NSF.

\bibliography{subtex/reference}
\bibliographystyle{plainnat}
\clearpage
\input{subtex/supplementary.tex}

\end{document}

%% file: subtex/intro.tex
\section{Introduction}
Deep learning has achieved state-of-the-art performance on a wide range of tasks, including image classification~\citep{krizhevsky2012imagenet}, language modeling~\citep{sutskever2014sequence}, and game playing~\citep{silver2016mastering}. One challenge in training deep neural networks (NNs) is that such models may overfit to the observed data, yielding over-confident decisions in learning tasks. This is partially because
most NN learning only seeks a point estimate for the model parameters, failing to quantify parameter uncertainty. 
A natural way to ameliorate these problems is to adopt a Bayesian neural network (BNN) formulation. By imposing priors on the weights, a BNN utilizes available data to infer an approximate posterior distribution on NN parameters~\citep{mackay1992practical}.  When making subsequent predictions (at test time), one performs model averaging over such learned uncertainty, effectively yielding a mixture of NN models~\citep{gal2016dropout,zhang2016towards,liu2016stein,li2016preconditioned,chen2017particle}. BNNs have shown improved performance on modern achitectures, including convolutional and recurrent networks~\citep{li2016learning,gan2016scalable,fortunato2017bayesian}.

For computational convenience, traditional BNN learning typically makes two assumptions on the weight distributions: independent isotropic Gaussian distributions as priors, and fully factorized Gaussian proposals as posterior approximation when adopting variational inference \citep{hernandez2015probabilistic,blundell2015weight,liu2016stein,feng2017learning,pu2017stein}. 
By examining this procedure, we note two limitations:
$(\RN{1})$ the independent Gaussian priors can ignore the anticipated structural information between the weights, 
and
$(\RN{2})$ the factorized Gaussian posteriors can lead to unreasonable approximation errors and underestimate model uncertainty (underestimate variances).

Recent attempts have been made to overcome these two issues. 
For example, \citep{louizos2016structured,sun2017learning} introduced
{\em structural priors} with the matrix variate Gaussian (MVG) distribution \citep{gupta1999matrix} to impose dependency between weights within each layer of a BNN. 
%
Further, non-parametric variational inference methods, {\it e.g.}, Stein variational gradient descent (SVGD) \citep{liu2016stein}, iteratively transport a set of particles to approximate the target posterior distribution (without making explicit assumptions about the form of the posterior, avoiding the aforementioned factorization assumption). SVGD represents the posterior approximately in terms of a set of particles (samples), and is endowed with guarantees on the approximation accuracy when the number of particles is exactly infinity \citep{liu2017stein_flow}. However, since the updates within SVGD learning involve kernel computation in the parameter space of interest, the algorithm can be computationally expensive in a high-dimensional space. This becomes even worse in the case of structural priors, where a large amount of additional parameters are introduced, rendering SVGD inefficient when directly applied for posterior inference.



We propose an efficient learning scheme for accurate posterior approximation of NN weights, adopting the MVG structural prior. 
We provide a new perspective to unify previous structural weight uncertainty methods~\citep{louizos2016structured,sun2017learning} via the Householder flow~\citep{tomczak2016improving}.
This perspective allows SVGD to approximate a target structural distribution in a lower-dimensional space, and thus is more efficient in inference. We call the proposed algorithm {\it Structural Stein Variational Gradient Descent} (S$^2$VGD). 

We investigate the use of our structural-weight-uncertainty framework for learning policies in sequential decision problems, including contextual bandits and reinforcement learning. 
In these models, uncertainty is particularly important because greater uncertainty on the weights typically introduces more variability into a decision made by a policy network~\citep{kolter2009near}, naturally leading the policy to explore. As more data are observed, the uncertainty decreases, allowing the decisions made by a policy network to become more deterministic as the environment is better understood (exploitation when the policy becomes more confident). In all these models, structural weight uncertainty is inferred by our proposed S$^2$VGD.

We conduct several experiments, first demonstrating that S$^2$VGD yields effective performance on classic classification/regression tasks. We then focus our experiments on the motivating applications, sequential decision problems, for which accounting for NN weight uncertainty is believed to be particularly beneficial. In these applications the proposed method demonstrates particular empirical value, while also being computationally practical. The results show that structural weight uncertainty gives better expressive power to describe uncertainty driving better exploration.

%


%% file: subtex/preliminary.tex
\section{Preliminaries}
\vspace{-3mm}
\subsection{Matrix variate Gaussian distributions}
The matrix variate Gaussian (MVG) distribution \citep{gupta1999matrix} has three parameters, describing the probability of a random matrix $\mathbf{W}\in\mathbb{R}^{\ell_1 \times \ell_2}$:
\vspace{-3mm}
{\small\begin{equation}
\begin{aligned}
p(\mathbf{W}) &\triangleq \mathcal{MN}(\mathbf{W};\mathbf{M,U,V})\\
&=\dfrac{ \exp \left(\dfrac{1}{2}\text{tr}[\mathbf{V}^{-1}(\mathbf{W-M})^{\top}\mathbf{U^{-1}(W-M)}] \right)}{(2\pi)^{\ell_1 \ell_2 /2}|\mathbf{V}|^{\ell_1 /2}|\mathbf{U}|^{\ell_2 /2}}
\end{aligned}
\end{equation}}
\vspace{-3mm}

\hspace{-0.0cm}where $\mathbf{M}\in \mathbb{R}^{\ell_1 \times \ell_2}$ is the mean of the distribution. $\mathbf{U} \in \mathbb{R}^{\ell_1\times \ell_1}$ and $\mathbf{V} \in \mathbb{R}^{\ell_2\times \ell_2}$ encode covariance information for the rows and columns of $\mathbf{W}$ respectively. The MVG is closely related to the multivariate Gaussian distribution.

\begin{lemma}[\cite{golub2012matrix}]\label{lem:mvg2gau}
Assume $\mathbf{W}$ follows the MVG distribution in (1), then
\begin{equation}
\vec(\mathbf{W})\sim \mathcal{N}(\vec(\mathbf{M}),\mathbf{V} \otimes \mathbf{U})
\end{equation}
where \vec($\mathbf{M}$) is the vectorization of $\mathbf{M}$ by stacking the columns of $\mathbf{M}$, and $\otimes$ denotes the standard Kronecker product~\citep{golub2012matrix}.
\end{lemma}

Furthermore, a linear transformation of an MVG distribution is still an MVG distribution.
\begin{lemma}[\cite{golub2012matrix}]\label{lem:mvg_transform}
Assume $\mathbf{W}$ follows the MVG distribution in (1), $\mathbf{A}\in\mathbb{R}^{\ell_2 \times \ell_1},\mathbf{C}\in\mathbb{R}^{\ell_2 \times \ell_1}$, then,
\begin{equation}
\begin{aligned}
\mathbf{B} &\triangleq \mathbf{AW}\sim \mathcal{MN}(\mathbf{B;AM,AUA^{\top},V})\\
\mathbf{B} &\triangleq \mathbf{WC}\sim \mathcal{MN}(\mathbf{B;MC,U,C^{\top} VC})
\end{aligned}
\end{equation}
\end{lemma}


\paragraph{MVG priors for BNNs}
For classification and regression tasks on data
$\mathcal{D} = \{\dv_1, \cdots, \dv_N\}$, where $\dv_i=\{ \mathbf{x}_i, \yv_i\}$,
with input $\mathbf{x}_i$ and output $\yv_i$, an $L$-layer NN parameterizes the mapping $\{g_{\ell}\}_{\ell=1}^L$, defining the prediction of $\yv_i$ for $\mathbf{x}_i$ as:
\begin{equation}\label{eq:nn}
\hat{\yv}_i = f (\xv_i)= g_{L} \circ g_{L-1} \circ \cdots \circ g_{0}(\mathbf{x}_i),~~\forall i.~
\end{equation}
where $\circ$ represents function composition, {\it i.e.}, $\mathcal{A} \circ \mathcal{B}$ means $\mathcal{A}$ is evaluated on the output of $\mathcal{B}$. Each layer $g_{\ell}$ represents a nonlinear transformation. For example, with the Rectified Linear Unit (ReLU) activation function \citep{nair2010rectified},
$g_{\ell}(\xv_i) = \ReLU(\mathbf{W}_{\ell}^{\top}\xv_i+\mathbf{b}_{\ell})$, where $\ReLU(x)\triangleq \max{(0,x)}$,
$\mathbf{W}_{\ell}$ is the weight matrix for the $\ell$th-layer, and $\mathbf{b}_{\ell}$ the corresponding bias term.

The MVG can be adopted as a prior for the weight matrix in each layer, to impose the prior belief that there are intra-layer weight correlations,
\begin{equation}~\label{mvg_layer}
\mathbf{W}_{\ell} \sim \mathcal{MN}(\mathbf{W};\mathbf{0,U_{\ell},V_{\ell}}),
\end{equation}
where the covariances $\mathbf{U}_{\ell}, \mathbf{V}_{\ell}$ have components drawn independently from $\InvGamma(a_0, b_0) $.
%
The parameters are $\thetav \triangleq\{ \mathbf{W}_{\ell}, \log \mathbf{U}_{\ell}, \log \mathbf{V}_{\ell}\}$, and the above distributions represent the prior $p(\thetav)$.
%
Our goal with BNNs is to learn layer-wise structured weight uncertainty, described by the posterior distribution
$p( \thetav |  \mathcal{D} ) \propto p( \thetav )p( \mathcal{D}  | \thetav )$, represented below as $p$ for simplicity.
When $\Umat= \sigma \Imat$ and $\Vmat= \sigma  \Imat$, we reduce to BNNs with independent isotropic Gaussian priors~\citep{blundell2015weight}.

\subsection{Stein Variational Gradient Descent}

SVGD considers a set of particles $\{\thetav_i\}_{i=1}^M$ drawn from distribution $q$, and transforms them to better match the target distribution $p$, by update:
\begin{equation}\label{eq:svgd}
\begin{aligned}
\thetav_i &\leftarrow \thetav_i + \epsilon \phi(\thetav_i),\\
\phi &= \arg\max_{\phi\in \mathcal{F}} \left\{\dfrac{\partial}{\partial \epsilon} \KL(q_{[\epsilon\phi]}||p)\right\},
\end{aligned}
\end{equation}
where  $q_{[\epsilon\phi]}$ is the updated empirical distribution, with $\epsilon$ as the step size, and $\phi$ as a function perturbation direction chosen to minimize the KL divergence between $q$
and $p$.
SVGD considers $\mathcal{F}$ as the unit ball of a vector-valued reproducing kernel Hilbert space (RKHS) $\mathcal{H}$ associated with a kernel $\kappa(\thetav,\thetav^{\prime})$. The RBF kernel is usually used as default. It has been shown~\citep{liu2016stein} that:
{\small\begin{align}\label{eq:ksd}
-\frac{\partial}{\partial \epsilon} \KL(q_{[\epsilon \phi]}\|p)|_{\epsilon=0} &= \mathbb{E}_{\thetav \sim q}[\Gamma_p \phi(\thetav)],\\
\text{with}
~~\Gamma_p \phi(\thetav) &\triangleq \nabla_{\thetav} \log p( \thetav |  \mathcal{D} ) ^{\top} \phi(\thetav) + \nabla_{\thetav} \cdot \phi(\thetav), \nonumber
\end{align}}
$\!\!\!$where $\nabla_{\thetav} \log p(\thetav)$ 
denotes the derivative of the log-density of $p$; $\Gamma_p$ is the Stein operator.
%
Assuming that the update function $\phi(\thetav)$ is in a RKHS with kernel $\kappa(\cdot,\cdot)$, it has been shown in~\citep{liu2016stein} that (\ref{eq:ksd}) is maximized with:
{\small
\begin{align}
\label{equ:close}
\phi(\thetav) = \mathbb{E}_{\thetav\sim q}[\kappa(\thetav, \thetav') \nabla_{\thetav} \log p( \thetav |  \mathcal{D} )
 + \nabla_{\thetav} \kappa(\thetav, \thetav')].
\end{align}}
$\!\!\!$The expectation $\mathbb{E}_{\thetav\sim q}[\cdot] $ can be approximated by an empirical averaging of particles $\{{\thetav_i}\}_{i=1}^M$, resulting in a practical SVGD procedure as:
\begin{align}  \label{eq:svgd_update}
\begin{aligned}
\thetav_i \leftarrow \thetav_i + \dfrac{\epsilon}{M} \sum_{j=1}^M \Big[ & \kappa(\thetav_j, \thetav_i) \nabla_{\thetav_j} \log p( \thetav_j |  \mathcal{D} )\\
  + & \nabla_{\thetav_j} \kappa(\thetav_j, \thetav_i) \Big].
\end{aligned}
\end{align}
%
The first term to the right of the summation in (\ref{eq:svgd_update}) drives the particles $\thetav_i$ towards the high probability regions of $p$, with information sharing across similar particles. The second term repels the particles away from each other, encouraging coverage of the entire distribution.
SVGD applies the updates in (\ref{eq:svgd_update}) repeatedly, and the samples move closer to the target distribution $p$ in each iteration. When using state-of-the-art stochastic gradient-based algorithms, {\it e.g.}, RMSProp~\citep{hinton2012rmsprop} or Adam~\citep{kingma2014adam}, SVGD becomes a highly efficient and scalable Bayesian inference method.

\paragraph{Computational Issues} Applying SVGD with structured distributions for NNs has many challenges.
For a weight matrix $\mathbf{W}$ of size $\ell_1 \times \ell_2$, the number of parameters in $\Umat$ and $\Vmat$ are $\ell_1^2$ and $\ell_2^2$, respectively.
Hence, the total number of parameters $\thetav$ needed to describe the distribution is $\ell_1 \ell_2+\ell_1^2+\ell_2^2$, compared to $\ell_1 \ell_2+1$ in traditional BNNs that employ isotropic Gaussian priors and factorization.

The  increase of parameter dimension by $\ell_1^2+\ell_2^2-1$ leads to significant computational overhead. The problem becomes even more severe in two aspects in calculating the kernels:
$(\RN{1})$ The computation increases quadratically by $M(M-1)(\ell_1^2+\ell_2^2-1)/2$,
$(\RN{2})$ the approximation to the repelling term in~\eqref{eq:svgd_update} using limited particles can be inaccurate in high dimensions.
Therefore, it is desirable to transform the MVG distribution to a lower-dimensional representation.


%% file: subtex/methods.tex
\vspace{-3mm}
\section{BNN \& Imposition of Structure}
\vspace{-3mm}
\subsection{Reparameterization of the MVG}
\vspace{-3mm}

Since the covariance matrices $\mathbf{U}$ and $\mathbf{V}$ are positive definite, we can decompose them as $
\mathbf{U} = \mathbf{P} \mathbf{\Lambda}_1 \mathbf{\Lambda}_1 \mathbf{P}^{\top}, \mathbf{V} = \mathbf{Q} \mathbf{\Lambda}_2 \mathbf{\Lambda}_2\mathbf{Q}^{\top}$, where
$\mathbf{P}$ and $\mathbf{Q}$ are the corresponding orthogonal matrices, $\mathbf{\Lambda}_1$ and $\mathbf{\Lambda}_2$ are diagonal matrices with positive diagonal elements.  
According to Lemma 2, we show the following reparameterization of MVG:
\begin{proposition}\label{pp:reparameterization}
	For a random matrix $\Cmat$ following an independent Gaussian distribution:  
	\begin{equation}
	\begin{aligned}
	\vec{ (\mathbf{C} ) } \sim \mathcal{N}( \cdot ,~\mathbf{P}^{\top} {\mathbf{\Lambda}}^{-1}_{1} \mathbf{M} \mathbf{\Lambda}^{-1}_{2} \mathbf{Q}, \mathbf{I} ),	
	\end{aligned}
	\end{equation}	
	the corresponding full-covariance MVG $\mathbf{W}$ in \eqref{lem:mvg2gau}  can be reparameterized as $\mathbf{W} = \mathbf{P} \mathbf{\Lambda}_1\mathbf{C} \mathbf{\Lambda}_2\mathbf{Q}^{\top}$. 
\end{proposition}
The proof is in Section A of the Supplementary Material.
Therefore, $\mathbf{W}$ drawn from MVG can be decomposed as the linear product of five random matrices: 
\begin{minipage}{1.00\linewidth}
	\begin{itemize}
		\vspace{1mm}
		\item 
		$\Cmat$ as a {\it standard weight matrix} with an independent Gaussian distribution.~\vspace{-2mm}
		\item 
		$\mathbf{\Lambda}_{1}$ and $\mathbf{\Lambda}_{2}$ as the {\it diagonal matrices}, encoding the structural information within each row and column, respectively.~\vspace{-2mm}
		\item 
		$\mathbf{P}$ and $\mathbf{Q}$ as {\it orthogonal matrices}, which characterize the structural information of weights between rows and columns, respectively.
	\end{itemize}
\end{minipage}

\subsection{MVG as Householder Flows}
\begin{figure}[t!] \centering
	\begin{tabular}{c}
		\hspace{-1mm}
		\includegraphics[width=8cm]{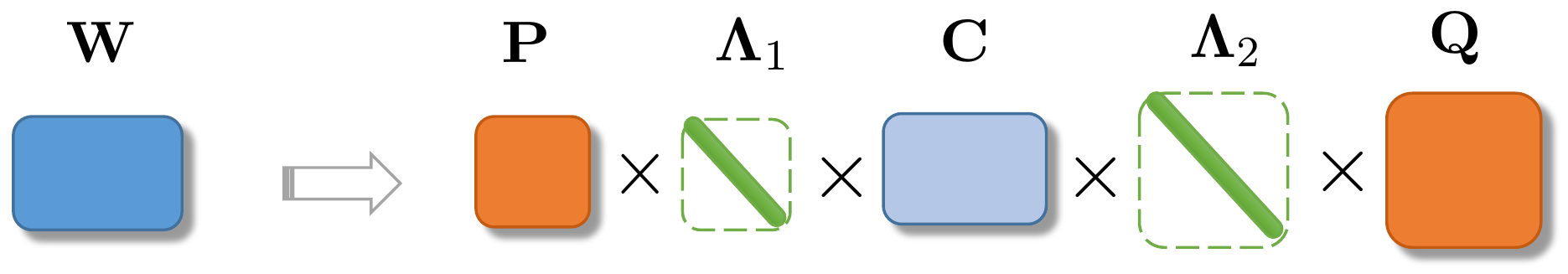}  
		\\
		(a) Layer Decomposition \vspace{2mm}
		\\
		\hspace{-1mm}
		\includegraphics[width=8cm]{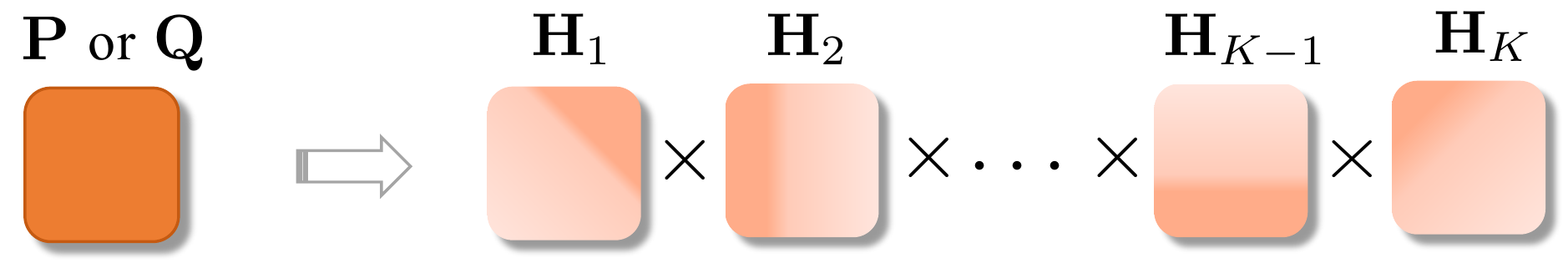}  
		\\
		(b) Householder Flow
	\end{tabular} \vspace{-4mm}
	\caption{\small Illustration of the two proposed techniques to reduce parameter size in learning the distribution of $\Wmat$: (a) decomposition of $\Wmat$ as a linear product of five matrices, and (b) approximation of $\Pmat$ or $\Qmat$ as a linear product of $K$ Householder matrices. Note that each rectangle indicates a matrix, ``$\times$'' indicates the matrix product, and $\Hmat$ in (b) is constructed using~\eqref{eq:householder_vec}.}
	\label{fig:shemes}
	\vspace{-4mm}
\end{figure}

Based on Proposition~\ref{pp:reparameterization}, we propose a {\it layer decomposition}: the one-layer weight matrix $\mathbf{W}$ with an MVG prior can be decomposed into a linear product of five matrices, as illustrated in Figure~\ref{fig:shemes}(a). Our layer decomposition provides an interesting interpretation for the original MVG layer: it is equivalent to several within-layer transformations. The representations imposed in the standard weight matrix $\Cmat$ are rotated by $\Pmat$ and $\Qmat$, and re-weighted by $\Lambdamat_1$ and $\Lambdamat_2$.

Note that the layer decomposition maintains similar computational complexity as the original MVG layer. To reduce the computation bottleneck in the layer decomposition, we further propose to represent $\Pmat$ and $\Qmat$ using {\it Householder flows}~\citep{tomczak2016improving}. 
Formally, a Householder transformation is a linear transformation that describes a reflection about a hyperplane containing the origin. Householder flow is a series of Householder transformations.

\begin{lemma}[\citep{sun1995basis}]\label{lem:householder}
	Any orthogonal matrix $\Mmat$ of degree K can be expressed as a Householder flow, \ie a product of {exactly} $K$ nontrivial Householder matrices., {\it i.e.},
	\begin{equation}
	\begin{aligned}
	\mathbf{M} = \Hmat_{K}\mathbf{H}_{K-1}\cdots\Hmat_{1}.
	\end{aligned}
	\end{equation}
\end{lemma}
Importantly, each Householder matrix $\Hmat$ is constructed from a Householder vector $\vv$ (which is orthogonal to the hyperplane):
\begin{equation}\label{eq:householder_vec}
\begin{aligned}
\Hmat = \Imat - 2\dfrac{\vv \vv^{\top}}{\|\vv\|^2}.
\end{aligned}
\end{equation}

According to Lemma~\ref{lem:householder}, $\Pmat$ and $\mathbf{Q}$ can be represented as a product of Householder matrices $\{\mathbf{H}^{(p)}_{k}\}$ and $\{\mathbf{H}^{(q)}_{k}\}$:

\begin{equation}
\begin{aligned}
& \mathbf{P} = \mathbf{H}^{(p)}_{K}\mathbf{H}^{(p)}_{K-1}\cdots\mathbf{H}^{(p)}_{1}\\
& \mathbf{Q} = \mathbf{H}^{(q)}_{K}\mathbf{H}^{(q)}_{K-1}\cdots\mathbf{H}^{(q)}_{1} \\
& \mathbf{H}^{(p)}_{k}  =\mathbf{I} - 2\vv^{(p)}_{k} \vv^{(p) \top }_{k}/ \left(\vv^{(p) \top }_{k} \vv^{(p)}_{k} \right)\\
& \mathbf{H}^{(q)}_{k}  =\mathbf{I} - 2 \vv^{(q)}_{k} \vv^{(q) \top }_{k}/ \left( \vv^{(q) \top }_{k} \vv^{(q)}_{k} \right),
\end{aligned}
\label{equ:Househoder}
\end{equation}
where $\vv^{(p)}_{k}$  and $\vv^{(q)}_{k}$ are the $k$th Householder vector for $\Pmat$ and $\Qmat$, respectively. This is illustrated in Figure~\ref{fig:shemes} (b). Note that the degree $K \leq \min \{\ell_1,\ell_2\} $, with proof in Section B of Supplementary Material.
In practice, $K$ is a trade-off hyperparameter, balancing the approximation accuracy and computation trade-off.

Since Householder flows allow one to represent $\Pmat$ or $\Qmat$ as $K$ Householder vectors, the parameter sizes reduce from $\ell_1^2$ and $\ell_2^2$ to $K \ell_1$ and $K \ell_2$. Overall, we can model $\Wmat$ with structured weight priors using only $(K+1)(\ell_1+\ell_2)+\ell_1\ell_2$ parameters.
Therefore, we can efficiently capture the structure information with only a slight increase of computational cost (\ie $(K+1)(\ell_1+\ell_2)$). 

Interestingly, our method provides a unifying perspective of previous methods on learning structured weight uncertainty. In terms of prior distributions, when $K=0$ (\ie $\Pmat=\Qmat=\Imat$), our reparameterization reduces to~\citep{louizos2016structured}. 
When $K=1$, and $\Lambdamat_1=\Lambdamat_2=\Imat$, our reparameterization reduces to \citep{sun2017learning}. In terms of posterior learning methods, when $\Pmat=\Qmat=\Lambdamat_1=\Lambdamat_2=\Imat$ and $M>1$, S$^2$VGD reduces to SVGD; when $M=1$, it reduces to learning an MAP solution.


\vspace{-3mm}
\subsection{Structural BNNs Revisited}
\vspace{-3mm}

We can leverage the layer decomposition and Householder flow above to construct an equivalent BNN by approximating the $\ell$th MVG layer in \eqref{mvg_layer} with standard Gaussian weight matrices:
\begin{equation}
\begin{aligned}
\label{equ:fullmode}
p(\mathbf{C}|\lambda) &= \mathcal{N}\left( \mathbf{C}_\ell ; \mathbf{0},\lambda \right),\\
p( \vv^{(p)}_{k\ell} |\phi) &=  \mathcal{N}\left(\vv^{(p)}_{k\ell} ;\mathbf{0}, \phi \mathbf{I} \right),\\
p(\vv^{(q)}_{k\ell} |\phi) &=  \mathcal{N}\left(\vv^{(q)}_{k\ell}  ;\mathbf{0}, \phi \mathbf{I} \right),\\
p(\mathbf{\Lambda}^{(1)} | \psi) & =  \mathcal{N}\left(\mathbf{\Lambda}_\ell^{(1)};\mathbf{0}, \psi \mathbf{I} \right)\\
p(\mathbf{\Lambda}^{(2)} | \psi) & = \mathcal{N}\left(\mathbf{\Lambda}_\ell^{(2)};\mathbf{0}, \psi  \mathbf{I} \right),\\
\lambda,  \phi, \psi &\sim \InvGamma ( \cdot;a_{\ell}, b_{\ell}).
\end{aligned}
\end{equation}
%
The forms of the likelihood for the last layer are defined according to the specific applications. For regression problems on real-valued response $\yv$:
\begin{equation}
\label{equ: regression}
\begin{aligned}
\mathbf{\mathbf{y}| \xv}, \Wmat_L  &\sim \mathcal{N}(\mathbf{y};f(\xv,\Wmat_L),\gamma\mathbf{I} )\\
\gamma &\sim \InvGamma ( \cdot;a_L, b_L)~,
\end{aligned}
\vspace{-2mm}
\end{equation}
with $f(\cdot)$ a neural network defined in \eqref{eq:nn}.
For classification problems on discrete labels $\yv$:
\begin{equation}
\label{equ: classifcation}
\begin{aligned}
\hspace{-0.2cm}\mathbf{\mathbf{y}| \xv} ,\Wmat_L &\sim \Categorical (\mathbf{y}; \Softmax (f ( \xv ,\Wmat_L ) ).
\end{aligned}
\end{equation}

Note that $\Wmat_L$ can follow the same proposed techniques to reduce parameter size.
Therefore, standard SVGD algorithms can be applied to sample from the posterior distribution of each model parameter. 
Intuitively, in SVGD the kernel function governs the interactions between particles, which employs this information to accelerate convergence and provide better performance. Similarly, the Householder flow, encoding structural information, controls the interactions between weights in each particle.


\section{Sequential Decision-Making}
A principal motivation for the proposed S$^2$VGD framework is sequential decision-making, including contextual multi-arm bandits (CMABs) and Markov decision processes (MDPs).   
A challenge in sequential decision-making in the face of uncertainty is the exploration/exploitation trade-off: the trade-off between either taking actions that are most rewarding according to the current knowledge, or taking exploratory actions,
which may be less immediately rewarding, but may lead to better-informed decisions in the future.
In a Bayesian setting, the exploration/exploration trade-off is naturally addressed by imposing uncertainty into the parameters of a policy model. 

\vspace{-2mm}
\subsection{CMABs and Stein Thompson Sampling}

CMABs model stochastic, discrete-time and finite action-state space control problems.
A CMAB is formally defined as a tuple $\mathcal{C} = \left \langle \Scal, \Acal, P_s, P_r, r \right \rangle$, where $\Scal$ is the state/context space, $\Acal$ the action/arm space, $r \in \mathbb{R}$ is the reward, $P_s$ and $P_r$ are the unknown environment distributions to draw the context and reward, respectively. At each time step $t$, the agent 
$(\RN{1})$ 
first observes a context $\sv_t \in \Scal$, drawn i.i.d. over time from $P_s$; then $(\RN{2})$  
chooses an action at $\av_t \in \Acal$ and observes a stochastic reward $r_t(\av_t, \sv_t)$, which is drawn i.i.d. over time from $P_r( \cdot |\av_t, \sv_t)$, conditioned on the current context and action.
%
%
%
The agent makes decisions via a policy $\pi(\av|\sv)$ that maps each context to a distribution over actions, yielding the probability of choosing action $\av$ in state $\sv$. The goal in CMABs is to learn a policy to maximize the expected total reward in $T$ interactions: 
$ J(\pi) = \E_{  P_s, \pi, P_r } \sum_{t=1}^T  r_t  $.


We represent the policy using a $\thetav$-parameterized neural network $\pi_{\thetav}(\av|\sv)$, where MVG priors $p(\thetav)$ are employed on the weights.
At each time $t$, given the past observations $\mathcal{D}_{t} \triangleq \{\dv\}_{j=1}^{t}$, where $\dv_j=(\sv_j, \av_j, r_j)$, the posterior distribution of $\thetav_t$ is updated as $p(\thetav_t|\mathcal{D}_t) \propto \prod_{j=1}^{t} p (\dv_j| \thetav) p(\thetav)$.

Thompson sampling~\citep{thompson1933likelihood} is a popular method to solve CMABs~\citep{li2011unbiased}. It approximates the posterior $p(\thetav|\mathcal{D}_t)$ in an online manner.
At each step, Thompson sampling 
$(\RN{1})$ first draws  a set of parameter samples, then 
$(\RN{2})$ picks the action by maximizing the expected reward over current step, {\it i.e.}, 
$\av_t = \arg\max_{\av} \mathbb{E}_{r \sim P_r (\cdot |\av, \sv_t; \thetav_t)} r_t$,  $(\RN{3})$ collects data samples after observing the reward $r_t$,
and $(\RN{4})$ updates posterior of the policy.
We apply the proposed S$^2$VGD for the updates in the final step, and call the new procedure Stein Thompson sampling, summarized in Algorithm~\ref{alg:Stein}. 

Note our Stein Thompson sampling is a general scheme for exploration/exploitation balance in CMABs. The techniques in \citep{russo2017time, kawale2015efficient} can be adapted in this framework; we leave this for future work.

\begin{algorithm}[htb]
	\caption{Stein Thompson Sampling}
	\label{alg:Stein}
	\begin{algorithmic}[1]
		\Require {$\mathcal{D} = \emptyset$; initialize particles $ \Theta_0  = \{\thetav_i\}_{i=1}^M$};
		\For{$t = 0,1,2,\ldots,T$}
		\State Receive context $\sv_t \sim P_s$;
		\State Draw a particle $\hat{\thetav}^{t}$ from $\Theta_t$;
		\State Select 
		$\av_t = \arg\max_{\av} \mathbb{E}_{r \sim P_r (\cdot |\av, \sv_t; \hat{\thetav}_t)} r_t$;
		\State Observe reward $r_t \sim P_r$, by performing $\av_t $;
		\State Collect observation: $\mathcal{D}_{t+1} = \mathcal{D}_t\cup(\sv_{t} , \av_{t}, r_{t})$; 
		\State Update $\Theta_{t+1}$, according to SVGD in (\ref{eq:svgd_update});
		\EndFor
	\end{algorithmic}
\end{algorithm}

\vspace{-2mm}
\subsection{MDPs and Stein Policy Gradient}
An MDP is a sequential decision-making procedure in a Markovian dynamical system. It can be seen as an extension of the CMAB, by replacing the context with the notion of a system state, that may dynamically change according to the performed actions and previous state. Formally, an MDP defines a tuple  $\mathcal{M} = \left \langle \Scal, \Acal, P_s, P_r, r, \gamma \right \rangle$, which is similar to a CMAB $\mathcal{C}$ except that $(\RN{1})$ the next state $\sv_{t+1}$ is now conditioned on state $\sv_t$ and action $\av_{t}$, \ie $\sv_{t+1} \sim P_s( \cdot| \sv_t,  \av_{t})$; and $(\RN{2})$ a discount factor $0 <\gamma <1$ for the reward is considered.
The goal is to find a policy $\pi(\av| \sv)$ to maximize the discounted expected reward: 
$ J(\pi) = \E_{  P_s, \pi, P_r } \sum_{t=1}^T \gamma^t  r_t$.




Policy gradient~\citep{sutton1998reinforcement} is a family of reinforcement learning methods that solves MDPs by iteratively updating the parameters $\thetav$ of the policy to maximize $J(\thetav) \triangleq J(\pi_{\thetav} (\av| \sv))$.  
Instead of searching for a single policy parameterized by $\thetav$, we consider
adopting an MVG prior for $p(\thetav)$, and learning its variational posterior distribution $q(\thetav)$ using S$^2$VGD. Following \citep{liu2017stein}, the objective function is modified as:  
\begin{equation}
\label{equ:rlobjective}
\begin{aligned}
\max_{q}\{\mathbb{E}_{q(\thetav)}[J(\thetav)] - \alpha \KL(q\|p)\},\vspace{-3mm}
\end{aligned}
\end{equation}
\vspace{-7mm}

where $\alpha \in [0, +\infty)$ is the temperature hyper-parameter to balance exploitation and exploration in the policy. The optimal distribution is shown to have a simple closed form~\citep{liu2017stein}:
\begin{equation}
\begin{aligned}
\label{equ: closedform}
q(\thetav)\propto \exp\left(\dfrac{1}{\alpha}J(\thetav)\right) p(\thetav).
\end{aligned}
\end{equation}
We iteratively approximate the target distribution as:
\begin{equation}
\begin{aligned}
\label{equ:SVGD}
\triangle \mathbf{\thetav}_i =\dfrac{\epsilon}{M} \sum_{j=1}^M [ &\nabla_{\mathbf{\thetav}_j}  \left(\dfrac{1}{\alpha}J(\thetav_j) + \log p(\thetav_j)\right) \kappa(\thetav_i, \thetav_j) \\
&+ \nabla_{\mathbf{\thetav}_j} \kappa(\mathbf{\thetav}_j, \mathbf{\thetav}_i) ],
\end{aligned}
\end{equation}
where $J(\thetav)$ can be approximated with REINFORCE~\citep{williams1992simple} or advantage actor critic~\citep{schulman2015high}.


We note two advantages of S$^2$VGD in sequential decision-making: (\RN{1}) the structural priors can characterize the flexible weight uncertainty, thus providing better exploration-exploitation when learning the policies; (\RN{2}) the efficient approximation scheme provides accurate representation of the true posterior while maintaining similar online-processing speed.

%% file: subtex/experiments.tex

\begin{table*}[t!]
	\centering
	\caption{Averaged predictions with standard deviations in terms of RMSE and log-likelihood on test sets.}
	\vskip 0.00in
	\hskip -0.02in
	\begin{adjustbox}{scale=0.80,tabular=c|cccc|cccc,center}
		\hline
		& \multicolumn{4}{c}{Test RMSE} &\multicolumn{4}{c}{Test Log likelihood} \\
		\hline
		Dataset & VMG &PBP\_MV & SVGD&S$^2$VGD & VMG  &  PBP\_MV &SVGD &S$^2$VGD\\
		\hline\hline
		Boston & 2.70 $\pm$ 0.13 & 2.76 $\pm$ 0.43 & 2.96$\pm$0.10&$\mathbf{ 2.56 \pm 0.33 }$ & -2.46 $\pm$ 0.09 & -3.01 $\pm$ 0.26 &-2.50$\pm$0.03&$\mathbf{ -2.43 \pm 0.10 }$\\
		Energy&0.54 $\pm$ 0.02&0.48 $\pm$ 0.04&1.37$\pm$0.05&$\mathbf{ 0.38 \pm 0.02 }$  &-1.06 $\pm$ 0.03&-2.37 $\pm$ 0.03&-1.77$\pm$0.02&$\mathbf{ -0.55 \pm 0.04 }$\\
		Concrete&4.89 $\pm$ 0.12 &4.66 $\pm$ 0.44&5.32$\pm$0.10&$\mathbf{ 4.25 \pm 0.37 }$ &-3.01 $\pm$ 0.03 &-3.22 $\pm$ 0.05&-3.08$\pm$0.02&$\mathbf{ -2.90 \pm 0.07 }$\\
		Kin8nm&0.08 $\pm$ 0.00&0.08 $\pm$ 0.00& 0.09$\pm 0.00$&$\mathbf{ 0.07 \pm 0.00 }$&1.10 $\pm$ 0.01&0.78 $\pm$ 0.02&0.98$\pm$0.01&$\mathbf{ 1.15 \pm 0.01 }$\\
		Naval&$\mathbf{0.00 \pm 0.00}$ &$\mathbf{0.00 \pm 0.00}$&$\mathbf{ 0.00 \pm 0.00 }$&$\mathbf{ 0.00 \pm 0.00 }$ &2.46 $\pm$ 0.00 &4.37 $\pm$ 0.17&4.09$\pm$0.01&$\mathbf{ 4.79 \pm 0.05 }$\\
		CCPP&4.04 $\pm$ 0.04&3.91 $\pm$ 0.09&4.03$\pm$0.03&$\mathbf{ 3.84 \pm 0.08 }$ &-2.82 $\pm$ 0.01 &-2.81 $\pm$ 0.02&-2.82$\pm0.01$&$\mathbf{ -2.77 \pm 0.02 }$\\
		Winequality &0.61 $\pm$ 0.04 &0.61 $\pm$ 0.02& 0.61$\pm$0.01&$\mathbf{0.59 \pm 0.02}$ &-0.95 $\pm$ 0.01 &-0.99 $\pm$ 0.07& -0.93$\pm$0.01&$\mathbf{-0.90 \pm 0.03}$ \\
		Yacht&0.48 $\pm$ 0.18 & 0.53 $\pm$ 0.14&0.86$\pm$0.05&$\mathbf{ 0.47 \pm 0.11 }$ &-1.30 $\pm$ 0.02 &-1.67 $\pm$ 0.24&-1.23$\pm$0.04&$\mathbf{ -0.81 \pm 0.14 }$\\
		Protein& $\mathbf{4.13 \pm 0.02}$ & 4.38 $\pm$ 0.01&4.61$\pm$0.01 & 4.15 $\pm$ 0.04 & $\mathbf{-2.84 \pm 0.00}$ & -2.91 $\pm$ 0.03& -2.95$\pm$0.00& -2.84 $\pm$ 0.01\\
		YearPredict&8.78 $\pm$ NA & 8.84 $\pm$ NA& $\mathbf{8.68\pm NA}$&8.73 $\pm$ NA &-3.59 $\pm$ NA &-3.58$\pm$NA&-3.62 $\pm$ NA & $\mathbf{ -3.57 \pm NA }$\\
		\hline
	\end{adjustbox}
	\label{tab:reg}
\end{table*}

\vspace{-4mm}
\section{Experiments}
\vspace{-3mm}
To demonstrate the effectiveness of our S$^2$VGD, we first conduct experiments on the standard regression and classification tasks, with real datasets (two synthetic experiments on classification and regression are given in the Supplementary Material). The superiority of S$^2$VGD  is further demonstrated in the experiments on contextual bandits and reinforcement learning.

We compare S$^2$VGD with related Bayesian learning algorithms, including VMG \citep{louizos2016structured}, PBP\_MV \citep{sun2017learning}, and SVGD \citep{liu2016stein}. The RMSprop optimizer is employed if there is no specific declaration.
For SVGD-based methods, we use a RBF kernel $\kappa(\thetav,\thetav') = \exp(-\|\thetav-\thetav'\|_2^2/h)$, with the bandwidth set to $h=\mathtt{med}^2/\log M$.~\citep{oates2016convergence, gorham2017measuring} Here $\mathtt{med}$ is the median of the pairwise distance between particles. The hyper-parameters $a_{\ell}=1, b_{\ell}=0.1$.
The experimental codes of this paper are available at: \href{https://github.com/zhangry868/S2VGD}{$\mathtt{https://github.com/zhangry868/S2VGD}$}.

%
We first study the role of hyperparameters in S$^2$VGD: the number of Householder transformations $K$ and the number of  particles $M$. This is investigated by a classification task from~\citep{liu2016stein} on the Covertype dataset with 581,012 data points and 54 features.

\begin{figure}[h!]
	\vspace{-4mm}
	\centering
	\hspace{-2mm}\includegraphics[width=6.5cm,height=4.2cm]{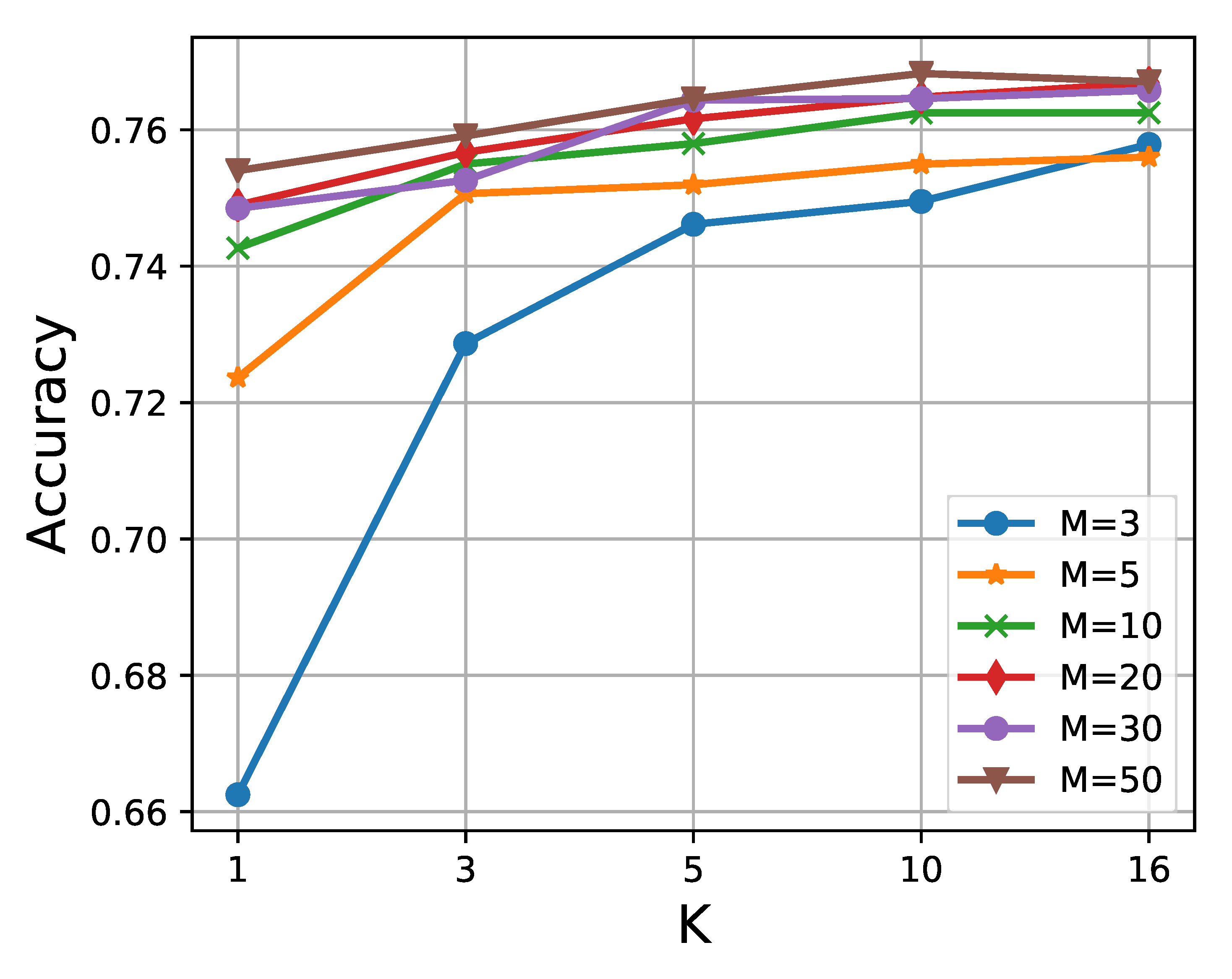}
	\vspace{-7mm}
	\caption{\small Impact of $K$ and $M$.}
	\label{fig:hyper}
	\vspace{-3mm}
\end{figure}

We perform 5 runs for each setting and report the mean of testing accuracy in Figure \ref{fig:hyper}. As expected, increasing $M$ or $K$ improves the performance, as they lead to a more accurate approximation. Interestingly, when $M$ is small, increasing $K$ gives significant improvement. Furthermore, when $M$ is large, the change of $K$ yields similar performance. Therefore, we set $K=1$ and $M=20$ unless otherwise specified.


	\vspace{-2mm}
	\subsection{Regression}
		\vspace{-2mm}
		
	We use a single-layer BNN for regression tasks. Following \citep{li2015stochastic},   10 UCI public datasets are considered: 100 hidden units for 2 large datasets (Protein and YearPredict), and 50 hidden units for the other 8 small datasets.
	%
 We repeat the experiments 20 times for all datasets except for Protein and YearPredict, which we repeat 5 times and once, respectively, for computation considerations~\citep{sun2017learning}.
 The batch size for the two large datasets is set to 1000, while it is 100 for the small datasets. The datasets are randomly split into 90\% training and 10\% testing.
 We adopt the root mean squared error (RMSE) and test log-likelihood as the evaluation criteria.
	
The experimental results are shown in Table~\ref{tab:reg}, from which we observe that $\RN{1})$ weight structure information is useful (SVGD is the only method without structure, and it yields inferior performance); $\RN{2}$) algorithms with non-parametric assumptions, \ie the Stein-based methods, perform better; and $\RN{3}$) when combined with structure information, our method achieves state-of-the-art results.
	
	\vspace{-2mm}
	\subsection{Classification}
		\vspace{-2mm}
	We perform the classification tasks on the standard MNIST dataset,
	which consists of handwritten digits of size $28\!\times \!28$, with 50,000 images for training and 10,000 for testing.
	A two-layer model 784-X-X-10 with $\ReLU$ activation function  is used, and X is the number of hidden units for each layer. The training epoch is set to 100. The test errors for network (X-X) sizes 400-400 and 800-800 are reported in Table \ref{tab:fnn}. We observe that the Bayesian methods generally perform better than their optimization counterparts. The proposed S$^2$VGD improves SVGD by a significant margin. Increasing $K$ also improves performance, demonstrating the advantages of incorporating structured weight uncertainty into the model. See \citep{li2016preconditioned,louizos2016structured,blundell2015weight} for details on the other methods with which we compare.
	
We wish to verify that the performance gain of S$^2$VGD is due to the special structural design of the network architecture, rather than the increasing number of model parameters. This is demonstrated by training a NN with 415-415 hidden units using SVGD, which yields test error $1.49\%$. It has slightly more parameters than our 400-400 network trained by S$^2$VGD (K=10), but worse performance.
	%

	\begin{table}[h]
		\centering
		\vskip -0.2in
		\centering
		\caption{Classification error of FNN on MNIST.}
		\vskip 0.1in
		\begin{adjustbox}{scale=1,tabular=llccc,center}
			\hline
			\multirow{2}{*}{Method } & \multicolumn{2}{c}{Test Error}  \\
			& 400-400 & 800-800 \\
			\hline
			S$^2$VGD (K=10)  & {\bf 1.36\%} & {\bf  1.30\%} \\
			S$^2$VGD (K=1)  & 1.43\% & 1.39\% \\
			SVGD  & 1.53\% &  1.47\% &   \\
			\hline
			SGLD	    	 &  1.64\% &    1.41\% \\
			RMSprop  	   &  1.59\% &    1.43\%  \\
			RMSspectral 	   		  &  1.65\% &    1.56\%  \\
			SGD 	   		  &   1.72\% &   1.47\%   \\
			\hline
			VMG, variational dropout   &  1.15\% & -  \\
			BPB, Gaussian   &  1.82\% & 1.99\%     \\
			BPB, scale mixture  &  {\bf 1.32}\% & 1.34\%  \\
			SGD, dropout  & 1.51\%  & 1.33\%  \\
			\hline
		\end{adjustbox}
		\label{tab:fnn}
		\vspace{-10pt}
	\end{table}
\begin{figure*}[t!] \centering
	\begin{tabular}{ccc}
		\hspace{-2mm}
		\includegraphics[width=5.6cm]{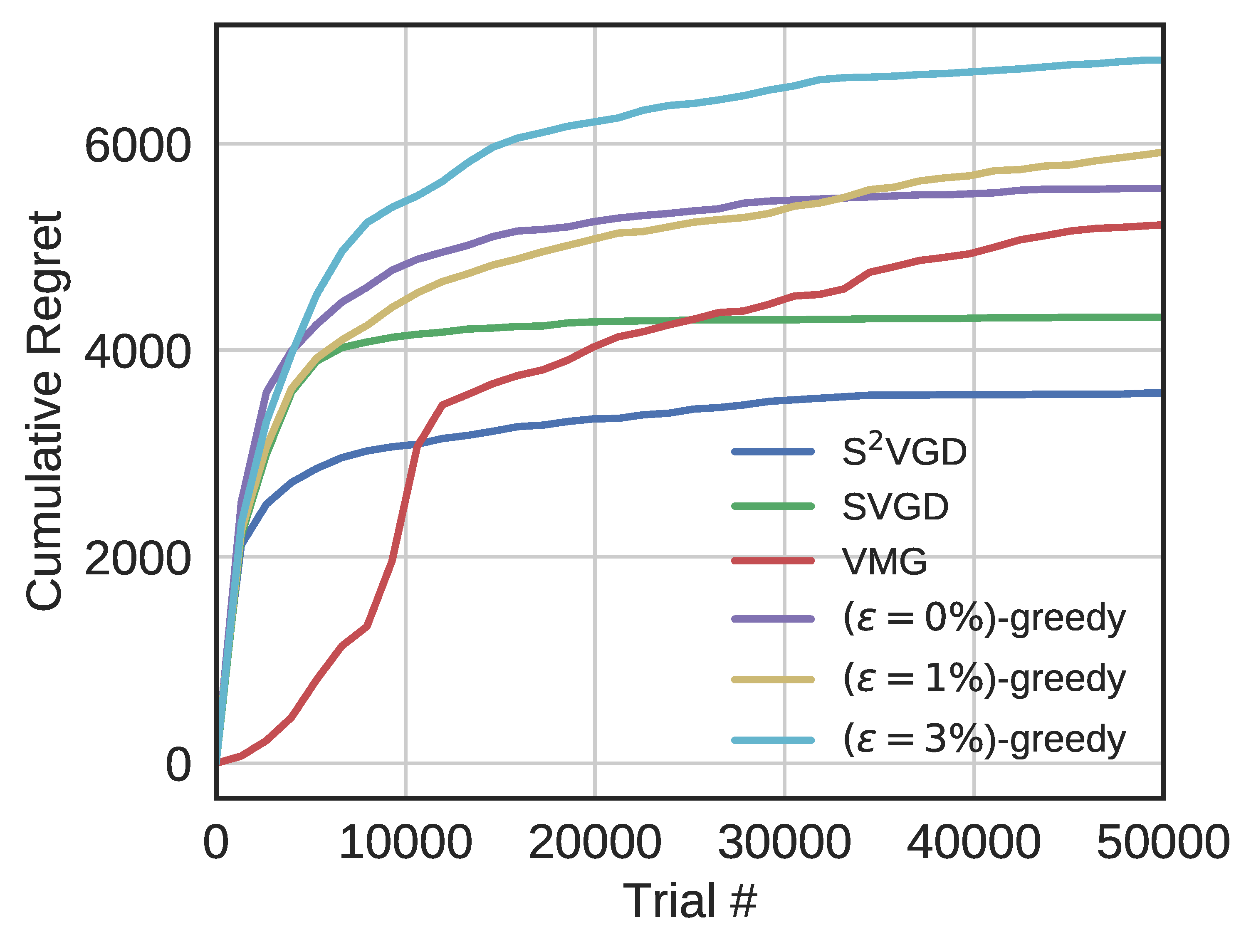}
		&   \hspace{-7mm}
		\includegraphics[width=5.5cm]{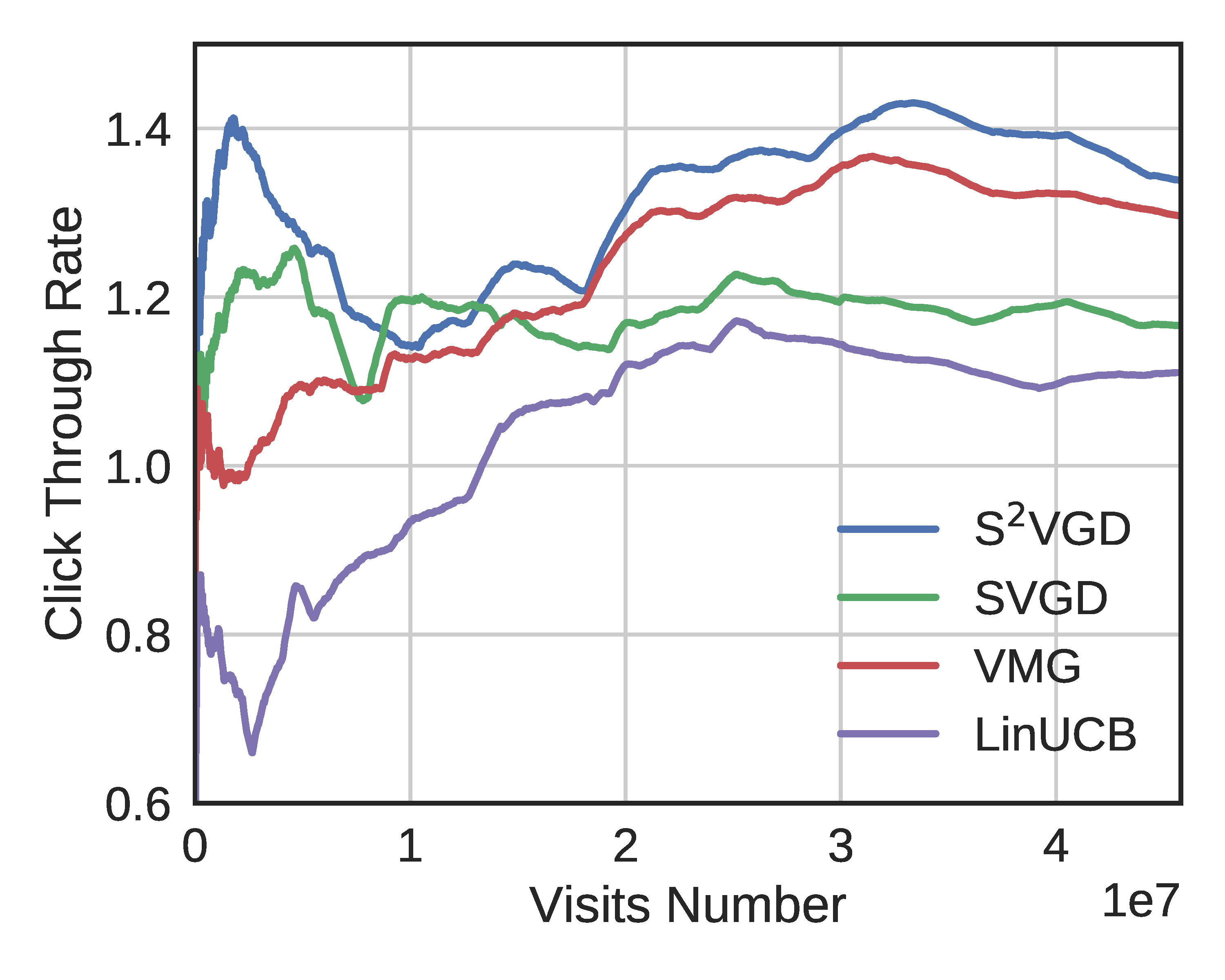}
		& 		\hspace{-5mm}
		\includegraphics[width=5.5cm]{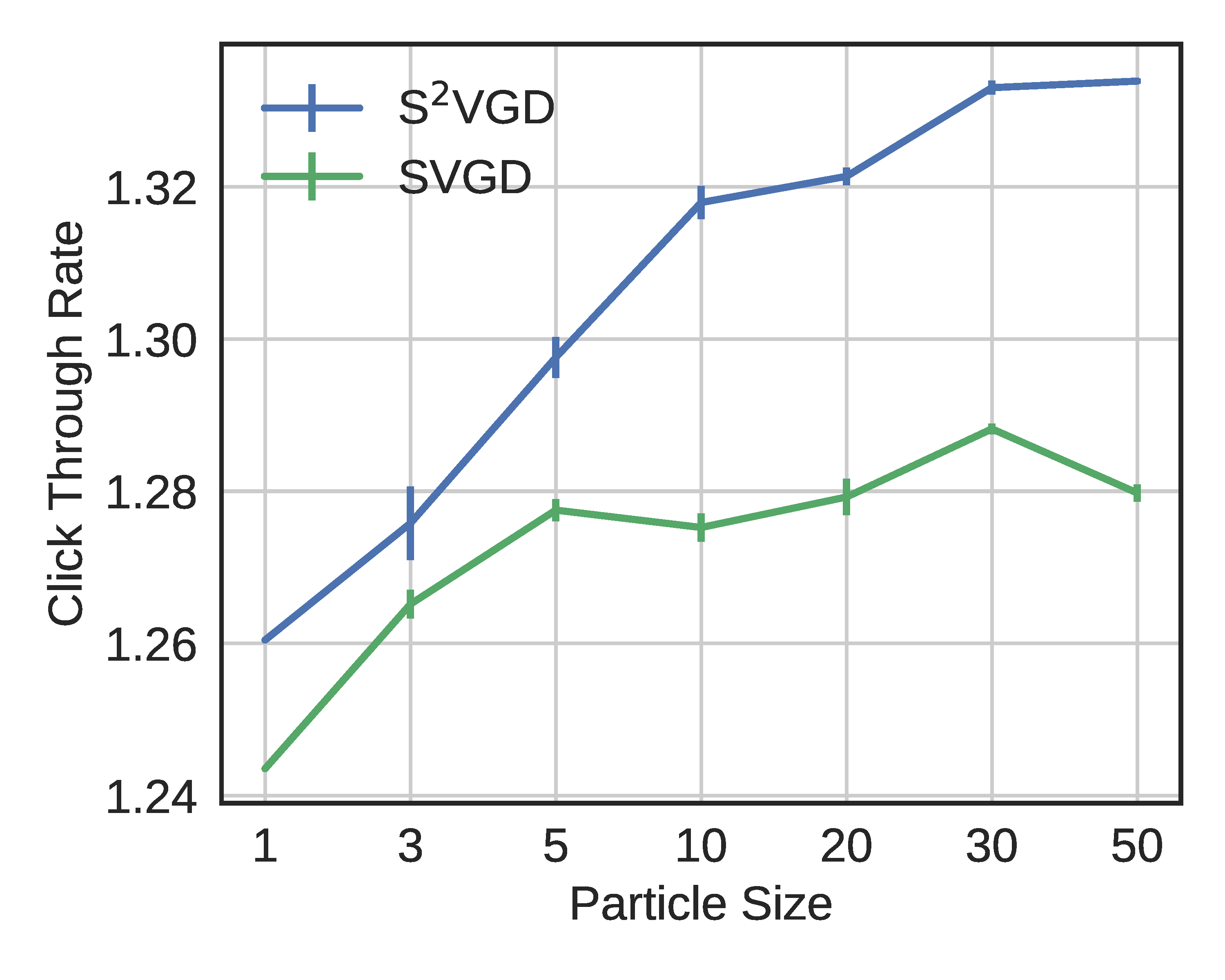}
		\vspace{-2mm}
		\\
		\hspace{-5mm}
		\small{(a) Simulation results on Mushroom} &
		\small{(b) New Article Recommendation} &
		\small{(c) Particle Size on Yahoo!Today}
	\end{tabular} \vspace{-2mm}
	\caption{{\small Experimental results of Contextual Bandits}}\label{fig:bandit}
	\vspace{-2mm}
\end{figure*}

 \vspace{-2mm}
\subsection{Contextual Bandits}
	\vspace{-2mm}
	
\paragraph{Simulation}
We first simulate a contextual-bandit problem with the UCI mushrooms dataset.  Following \citep{blundell2015weight}, the provided features of each mushroom are regarded as the context. A reward of 5 is given when an agent eats an edible mushroom. Otherwise, if a mushroom is poisonous and the agent eats it, a reward of -10 or 5 will be received, both with probability 0.5; if the agent decides not to eat the mushroom, it receives a reward of 0. We use a two-layer BNN with $\ReLU$ and 50 hidden units to represent the policy.
We compared our method with a standard baseline,
$\varepsilon$-greedy policy with $\varepsilon = 0\%$ (pure greedy), $1\%, 3\%$, respectively~\citep{sutton1998reinforcement}.

%
%
We evaluate the performance of different algorithm by cumulative regret~\citep{sutton1998reinforcement}, a measure of the loss caused by playing suboptimal bandit arms. The results are plotted in Figure~\ref{fig:bandit}(a).
%
%
Thompson sampling with 3 different strategies to update policy are considered: S$^2$VGD, SVGD and VMG.
S$^2$VGD shows lower regret at the beginning of learning than SVGD, and the lowest final cumulative regret among all methods. We hypothesize that our method captures the internal weight correlation, and the structural uncertainty can effectively help the agent learn to make less mistakes in exploration with less observations.


\vspace{-4mm}
\paragraph{News Article Recommendation}
We consider personalized news article recommendation on Yahoo! \citep{li2010contextual}, where each time a user visits the portal, a news article from a dynamic pool of candidates is recommended based on the user's profile (context). The dataset contains 45,811,883 user visits to the Today Module in a 10-day period in May 2009. For each visit, both the user and each of the 20 candidate articles are associated with a feature vector of 6 dimensions \citep{li2010contextual}.

The goal is to recommend an article to a user based on its behavior, or, formally, maximize the total number of clicks on the recommended articles. The procedure is regraded as a CMAB problem, where articles are treated as arms. The reward is defined to be 1 if the article is clicked on and 0 otherwise. A one-layer NN with $\ReLU$  and 50 hidden units is used as the policy network. The classic LinUCB~\citep{li2010contextual} is also used as the baseline.   The performance is evaluated by  an unbiased offline evaluation protocol: the average normalized accumulated click-through-rate (CTR) in every 20000 observations~\citep{li2010contextual,li2011unbiased}. The normalized CTRs are plotted in Figure~\ref{fig:bandit}(b). It is clear that S$^2$VGD consistently outperforms other methods. The fact that S$^2$VGD and VMG perform better than SVGD and the baseline LinUCB indicates that structural information helps algorithms to better balance exploration and exploitation.

To further verify the influence of the particle size $M$ on the sequential decision-making, we vary the $M$ from 1 to 50 on the-first-day data. All algorithms are repeated 10 times and their mean performances are plotted in Figure~\ref{fig:bandit}(c). We observe that the CTR keeps improving when $M$ becomes larger. S$^2$VGD dominates the performance of SVGD with much higher CTRs,  the gap becomes larger as $M$ increases. Since larger $M$ typically leads to more accurate posterior estimation of the policy, indicating again that the accurately learned structural uncertainty are beneficial for CMABs.


\subsection{Reinforcement Learning}
%
We apply our S$^2$VGD to policy gradient learning.
All experiments are conducted with the OpenAI $\mathtt{rllab}$ toolkit~\citep{duan2016benchmarking}. Three classical continuous control tasks are considered: Cartpole Swing-Up, Double Pendulum, and Cartpole. Following the settings in \citep{liu2017stein,houthooft2016vime},
the  policy is parameterized as a two-layer (25-10 hidden units) neural network with $\mathtt{tanh}$ as the activation function.
The maximal length of horizon is set to 500. SVGD and S$^2$VGD use a sample size of 10000 for policy gradient estimation, and $M=16$.
For the easy task, Cartpole, all agents are trained for 100 episodes. For the two complex tasks, Cartpole Swing-Up and Double Pendulum, all agents are trained up to 1000 episodes. We consider two different methods to estimate the gradients:  REINFORCE~\citep{williams1992simple} and advantage actor critic~(A2C)~\citep{schulman2015high}

\begin{figure}[t!] \centering
	\begin{tabular}{cc}
		\hspace{-5mm}
		\includegraphics[width=4.3cm]{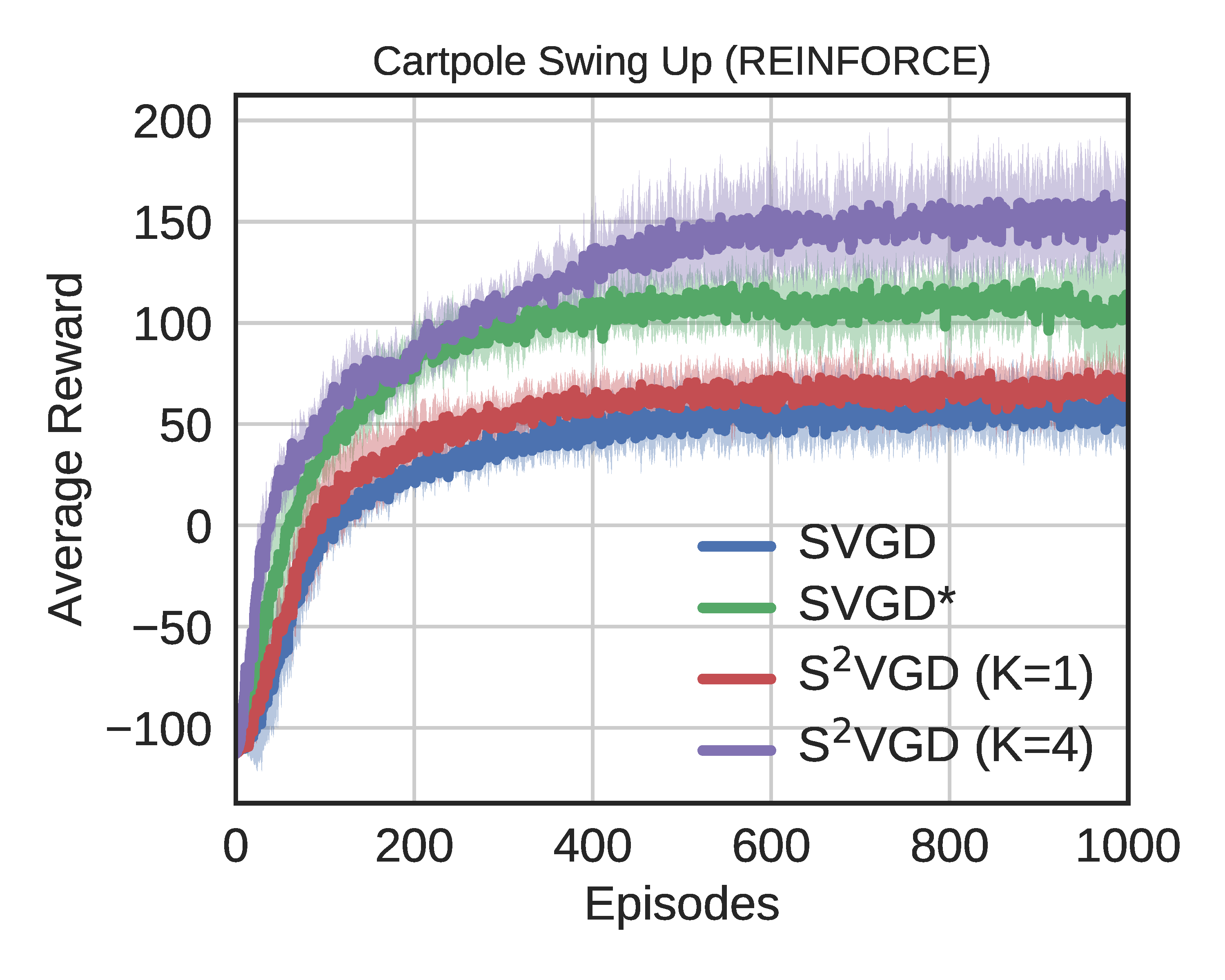}  &
		\hspace{-6mm}
		\includegraphics[width=4.3cm]{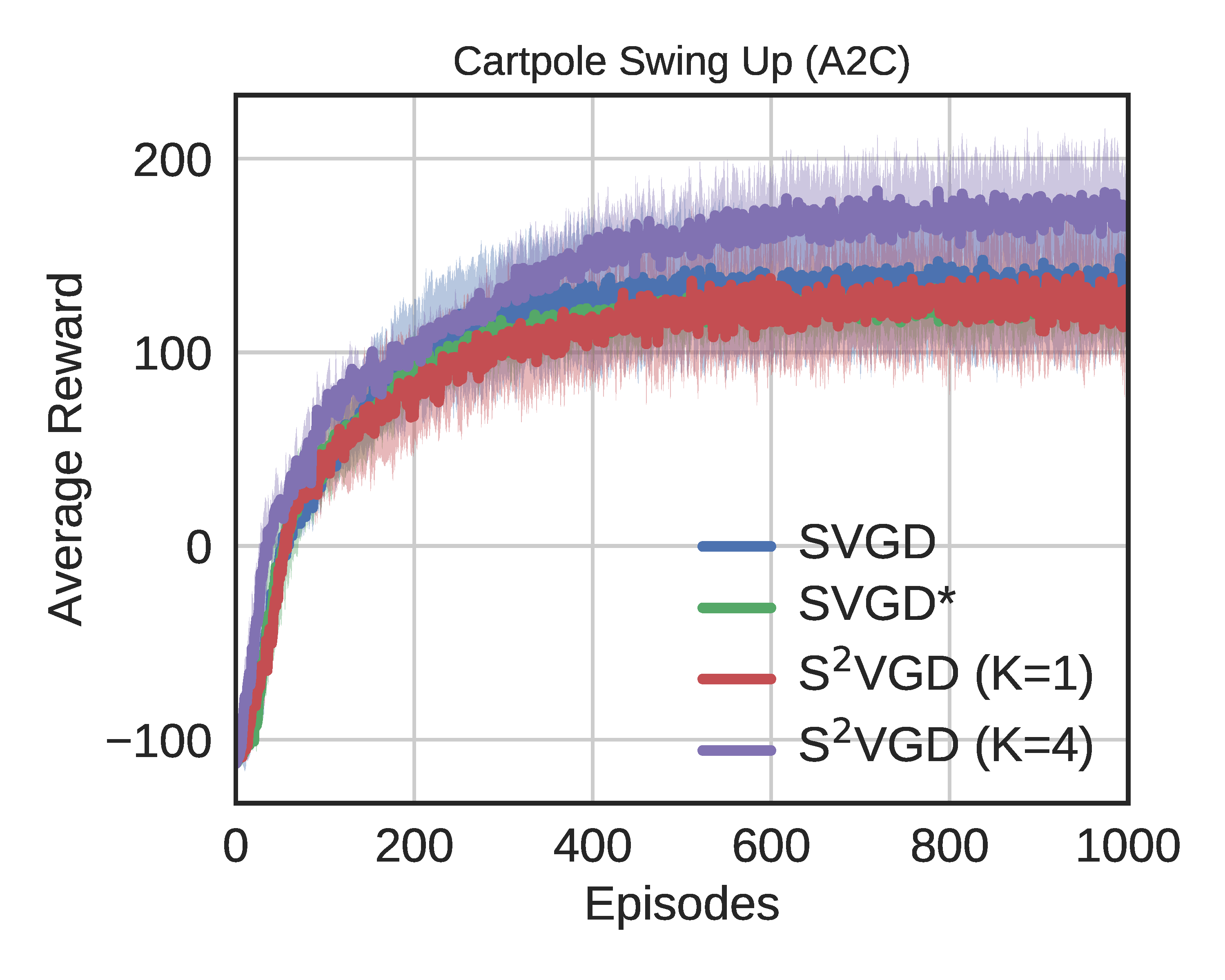}
		\vspace{-2mm}
		\\
		\hspace{-5mm}
		\includegraphics[width=4.3cm]{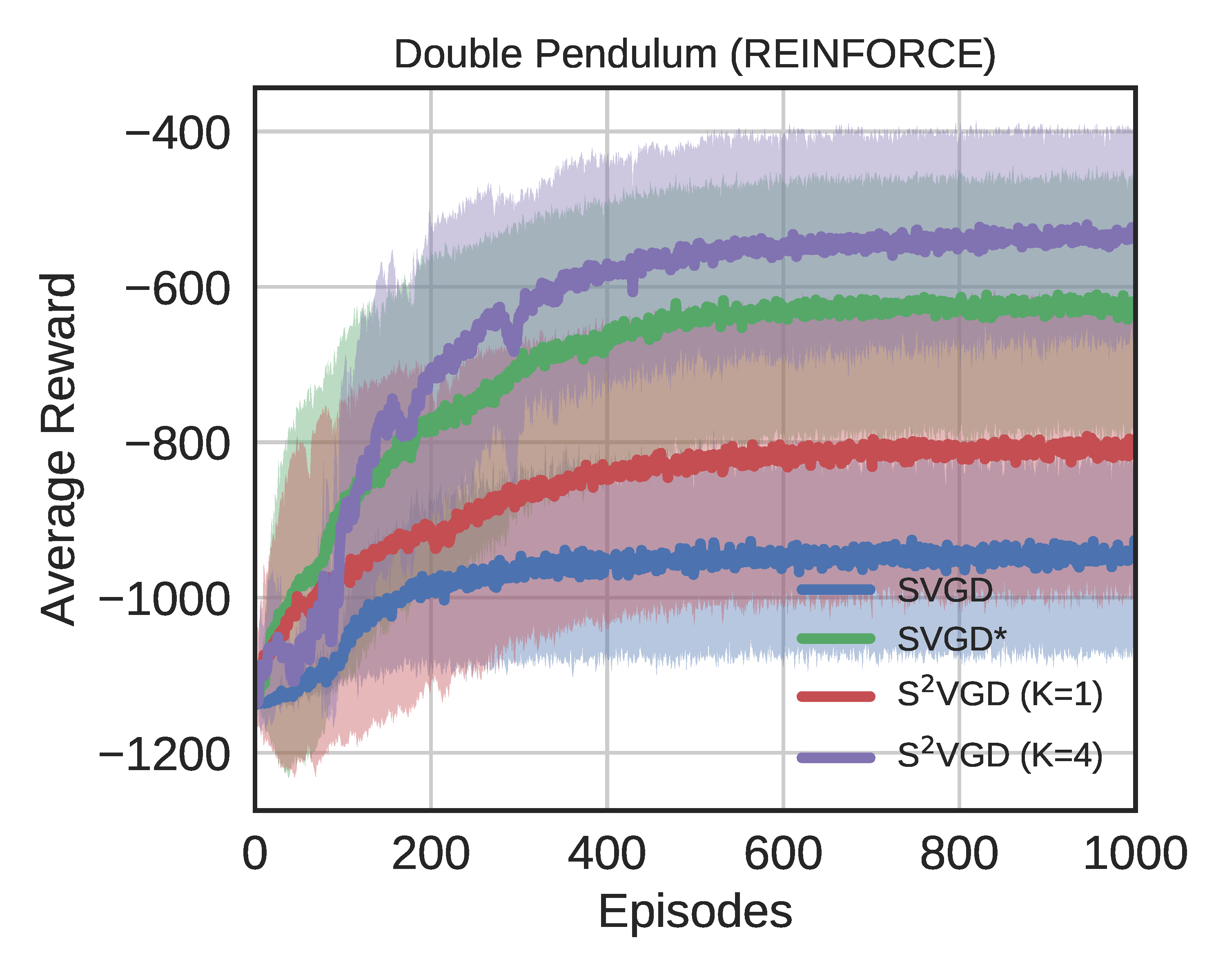} &
		\hspace{-6mm}
		\includegraphics[width=4.3cm]{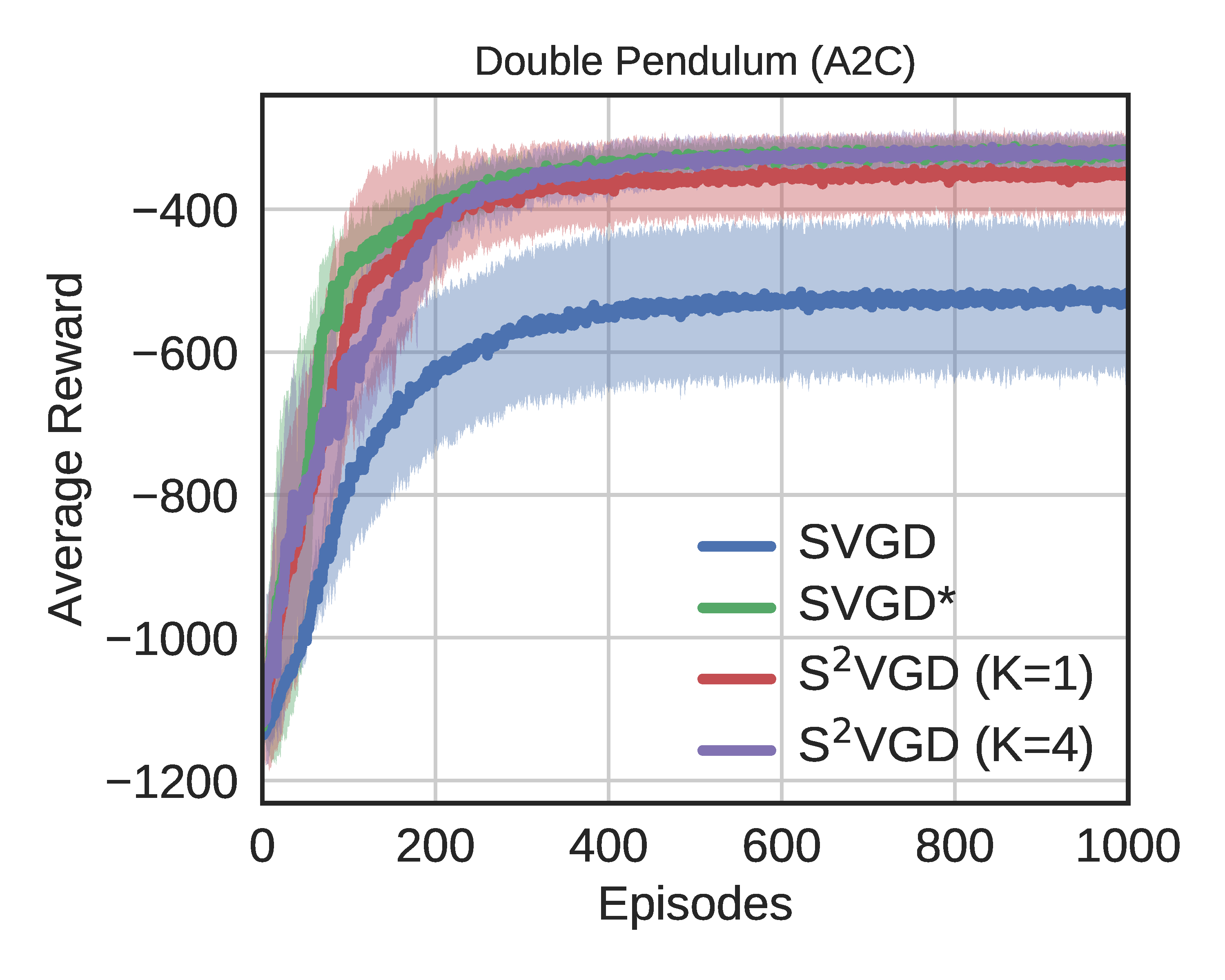}
		\vspace{-2mm}
		\\
		\hspace{-5mm}
		\includegraphics[width=4.3cm]{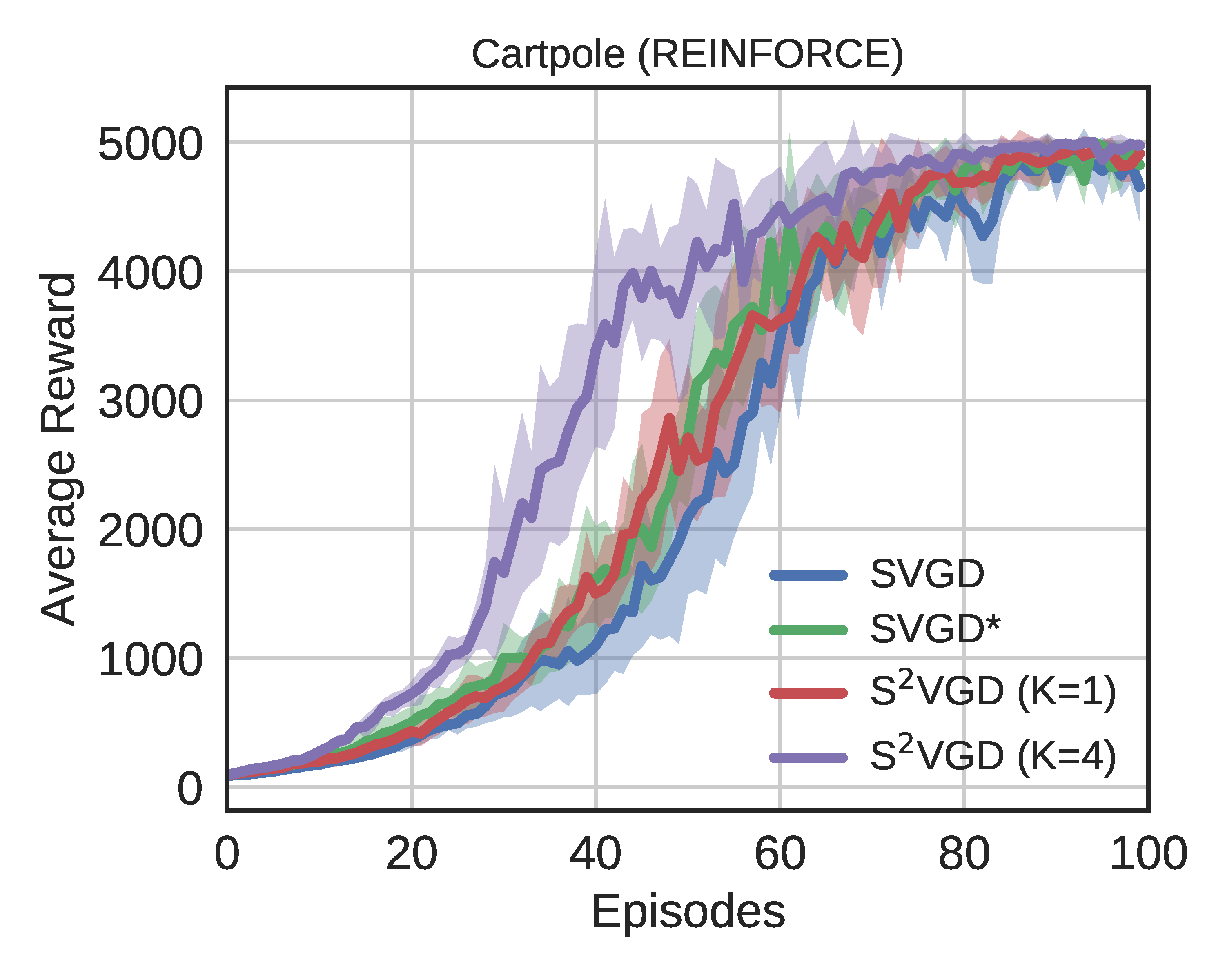}   &
		\hspace{-6mm}
		\includegraphics[width=4.3cm]{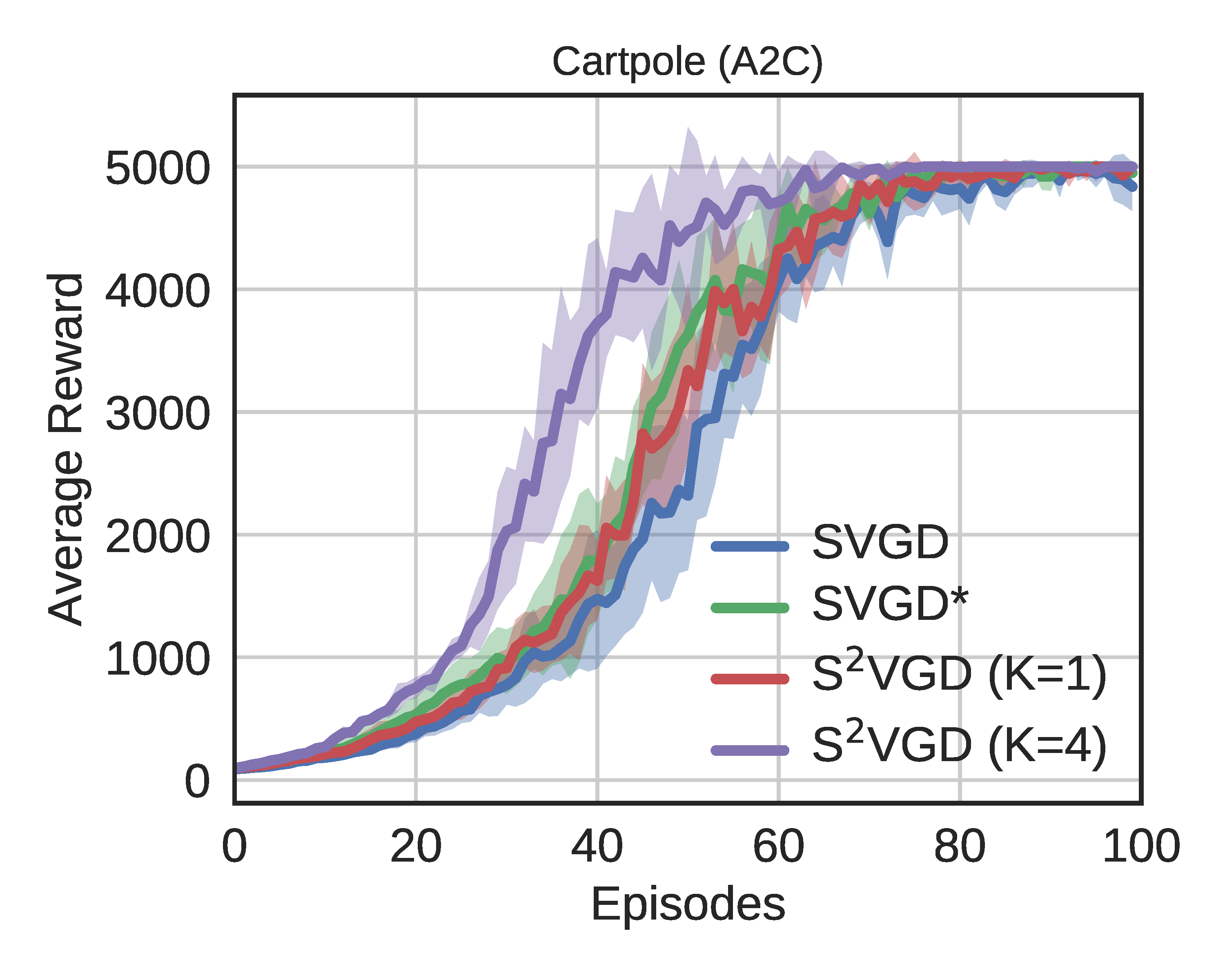}
		\\
		\end{tabular} \vspace{-7mm}
	\caption{{\small Learning curves by S$^2$VGD and SVGD with REINFORCE (left) and A2C (right).}}
	\vspace{-6mm}
	\label{fig:rlSVGD}
\end{figure}

Figure~\ref{fig:rlSVGD} plots the mean (dark curves) and standard derivation (light areas) of discounted rewards over 5 runs.
In all tasks and value-estimation setups, S$^2$VGD converges faster than SVGD and finally converges to higher average rewards.
The results are even comparable to~\citep{houthooft2016vime}, in which a subtle reward mechanism is incorporated to encourage exploration. It demonstrates that simply adding structural information on policy networks using S$^2$VGD improves the agent's exploration ability. We also add a baseline method called SVGD* that applies SVGD to train a network of similar size (25-16 hidden units) with the one reparameterized by S$^2$VGD ($K$=4).
The fact that S$^2$VGD ($K$=4) converges better than SVGD* demonstrates that our structural uncertainty is key to excellent performance.

%% file: subtex/conclusion.tex
	\vspace{-2mm}
\section{Conclusions}
\vspace{-2mm}
We have proposed S$^2$VGD, an efficient Bayesian posterior learning scheme for the weights of BNNs with structural MVG priors. 
To achieve this, we derive a new reparametrization for the MVG to unify previous structural priors, and adopt the SVGD algorithm for accurate posterior learning. By transforming the MVG into a lower-dimensional representation, S$^2$VGD avoids computation of related kernel matrices in high-dimensional space. 
The effectiveness of our framework is validated on several real-world tasks, including regression, classification, contextual bandits and reinforcement learning. Extensive experimental results demonstrate its superiority relative to related algorithms.

Our empirical results on sequential decision-making suggest the benefits of including inter-weight structure within the model, when computing policy uncertainty for online decision-making in an uncertain environments. More sophisticated methods for leveraging uncertainty for exploration/exploration balance may be a promising direction for future work. For example, explicitly encouraging exploration using learned structural uncertainty~\citep{houthooft2016vime}.



%% file: subtex/supplementary.tex
\medskip
\twocolumn[
\aistatstitle{\Large Learning Structural Weight Uncertainty for \\Sequential Decision-Making:
	Supplementary Material}

\aistatsauthor{  Ruiyi Zhang$^{1}$~~ Chunyuan Li$^{1}$~~  Changyou Chen$^{2}$~~ Lawrence Carin$^{1}$}

\aistatsaddress{ $^{1}$Duke University~~ $^{2}$University at Buffalo \\
	\texttt{ryzhang@cs.duke.edu,~cl319@duke.edu,~cchangyou@gmail.com,~lcarin@duke.edu}  }

]

\appendix

\section{Proof of Proposition 3}
\begin{proof}
	Let a MVG distributed matrix $\Wmat$ be
	\begin{equation}~\label{mvg_layer_supp}
	\mathbf{W} \sim \mathcal{MN}(\mathbf{W};\mathbf{M,U,V}),
	\end{equation}
	
	Since the covariance matrices $\mathbf{U}$ and $\mathbf{V}$ are positive definite, we can decompose them as
	\begin{align}
	\mathbf{U} &= \mathbf{P} \mathbf{\Lambda}_1 \mathbf{\Lambda}_1 \mathbf{P}^{\top},\\ \mathbf{V} &= \mathbf{Q} \mathbf{\Lambda}_2 \mathbf{\Lambda}_2\mathbf{Q}^{\top},
	\end{align}
	where $\mathbf{P}$ and $\mathbf{Q}$ are the corresponding orthogonal matrices, {\it i.e.}, $\Pmat\Pmat^{\top}=\Imat$, $\Qmat\Qmat^{\top}=\Imat$.
	
	According to Lemma \ref{lem:mvg_transform}, we have,
	\begin{equation}~\label{}
	\Pmat^{\top}\mathbf{W} \sim \mathcal{MN}(\Pmat^{\top}\mathbf{W};\Mmat,\Pmat^{\top}\Umat\Pmat,\Vmat),
	\end{equation}
	
	Since $\Umat = \mathbf{P} \mathbf{\Lambda}_1 \mathbf{\Lambda}_1 \mathbf{P}^{\top}$, and we have:
	\begin{equation}~\label{}
	\Pmat^{\top}\mathbf{W} \sim \mathcal{MN}(\Pmat^{\top}\mathbf{W};\mathbf{M},\Pmat^{\top} \mathbf{P} \mathbf{\Lambda}_1 \mathbf{\Lambda}_1 \mathbf{P}^{\top}\Pmat,\Vmat).
	\end{equation}
	Then,
	\begin{equation}~\label{}
	\Pmat^{\top}\mathbf{W} \sim \mathcal{MN}(\Pmat^{\top}\mathbf{W};\mathbf{M}, \mathbf{\Lambda}_1 \mathbf{\Lambda}_1 ,\Vmat),
	\end{equation}
	Similarly, we have,
	\begin{equation}~\label{}
	\Pmat^{\top}\mathbf{W}\Qmat \sim \mathcal{MN}(\Pmat^{\top}\mathbf{W}\Qmat;\mathbf{M}, \mathbf{\Lambda}_1 \mathbf{\Lambda}_1 , \mathbf{\Lambda}_2 \mathbf{\Lambda}_2).
	\end{equation}
	Further,
	{\small
		\begin{equation}~\label{eq:c}
		\mathbf{\Lambda}_{1}^{-1}\Pmat^{\top}\mathbf{W}\Qmat\mathbf{\Lambda}_{2}^{-1} \sim \mathcal{MN}(\mathbf{\Lambda}_{1}^{-1}\Pmat^{\top}\mathbf{W}\Qmat\mathbf{\Lambda}_{2}^{-1};\mathbf{0}, \Imat,\Imat),
		\end{equation}
	}
	$\!\!\!$Define $\Cmat = \mathbf{\Lambda}_{1}^{-1}\Pmat^{\top}\mathbf{W}\Qmat\mathbf{\Lambda}_{2}^{-1}$, then $\Cmat$ follows an independent Gaussian distribution:
	\begin{equation}
	\begin{aligned}
	\label{eq:indMVGGaussian}
	\Cmat \sim \mathcal{MN}(\Cmat;\mathbf{P}^{\top} {\mathbf{\Lambda}}^{-1}_{1} \mathbf{M} \mathbf{\Lambda}^{-1}_{2} \mathbf{Q}, \Imat,\Imat),
	\end{aligned}
	\end{equation}
	(\ref{eq:indMVGGaussian}) can also be expressed as:
	\begin{align}
	\vec{ (\mathbf{C} ) } \sim \mathcal{N}(\vec{ (\mathbf{C}) };\mathbf{P}^{\top} {\mathbf{\Lambda}}^{-1}_{1} \mathbf{M} \mathbf{\Lambda}^{-1}_{2} \mathbf{Q}, \mathbf{I} ),	
	\end{align}
	showing that vectorized elements form of $\Cmat$ follows an isotropic Gaussian distribution.
	Finally, since $\mathbf{\Lambda}_{1}\Cmat\mathbf{\Lambda}_{2} $=$ \Pmat^{\top}\mathbf{W}\Qmat$ , we have:
	\begin{align}
	\mathbf{W}=\Pmat\mathbf{\Lambda}_{1}\Cmat\mathbf{\Lambda}_{2}\Qmat^{\top}~.
	\end{align}
\end{proof}

\section{Properties of Householder Flow}

\subsection{The Upperbound of Orthogonal Degree}
\begin{lemma}[\citep{sun1995basis}~The Basis-Kernel Representation of Orthogonal Matrices]
For any $m\times m$ orthogonal matrix $\Qmat$, there exist a full-rank $m\times k$ matrix $\Ymat$ and a nonsingular $k\times k$ matrix $\Smat$, $k\leq m$, such that:
\begin{equation}
	\Qmat \triangleq \Qmat(\Ymat,\Smat)=\Imat-\Ymat\Smat\Ymat^{\top}
\end{equation}	
\end{lemma}

\begin{definition}[\citep{sun1995basis}~Active Subspace]
Orthogonal matrix $\Qmat$ acts on the space $\mathcal{R}(\Ymat)^{\perp}$ as the identity and changes every nonzero vector in $\mathcal{R}(\Ymat)$, and $\mathcal{R}(\Ymat)^{\perp}$ is the active space of $\Qmat$
\end{definition}

In basis-kernel representation, $\Smat$ is the kernel, and $\Ymat$ is the basis.  The {\it degree of an orthogonal matrix} is defined as the dimension of its active subspace. Specifically, Householder matrix is an orthogonal matrix of degree 1.
With the introduced definitions and Lemma 5, the degree of an orthogonal matrix is bounded by Lemma 7.

\begin{lemma}[\citep{sun1995basis}]\label{lem:Kbound}
	Let $\mathbf{A}$ and $\mathbf{B}$ be two $m$-by-$k$ matrices, $k < m$. If $\mathbf{B} = \mathbf{QA}$ for some orthogonal matrix $\mathbf{Q}$, then $\mathbf{Q}$ is either of degree no greater than $k$ or can be replaced by an orthogonal factor of its own with degree no greater than $k$.
\end{lemma}

Since the degree of orthogonal matrix is bounded  by size of the matrix ($k$), the number of Householder transformations needed for Householder flow is also bounded.

\subsection{Intuitive Explanation}
In our setting, the orthogonal matrix works as a rotation matrix. The Householder transformation reflects the weights by a hyperplane orthogonal to its corresponding Householder vector. Hence, Householder flows applies a series of reflections to the original weights. In another word, it rotates the weight matrix, equivalent to the effects of the rotation matrix.

%

\section{Computational Trick}
Assume $\vv$ is a Householder vector, and its corresponding Householder matrix $\Hmat \triangleq \Imat - 2\dfrac{\vv \vv^{\top}}{\|\vv\|^2}$.  For a feature vector $\xv$, defining $\hat{\vv}\triangleq\vv/\|\vv\|$, then it is easy to show:
\begin{align}\label{eq:effhouseholder}
\Hmat\xv &=   \xv - 2\dfrac{\vv \vv^{\top}\xv}{\|\vv\|^2}\\
&=\xv-2\vv^{\top}\xv \dfrac{\vv}{\|\vv\|^2}\\
&= \xv - 2\langle\xv,\hat{\vv}\rangle \hat{\vv}~.
\end{align}

To efficiently compute the Householder transformation, we do not need to revert the Householder matrix $\Hmat$ and apply $\Hmat \xv$ to complete one Householder transformation. (\ref{eq:effhouseholder}) can be used to drastically reduce the computational cost and thus make the Householder flows more efficient.

\section{Experimental Results}
\begin{figure}[ht]\centering
	\includegraphics[width=0.85\linewidth]{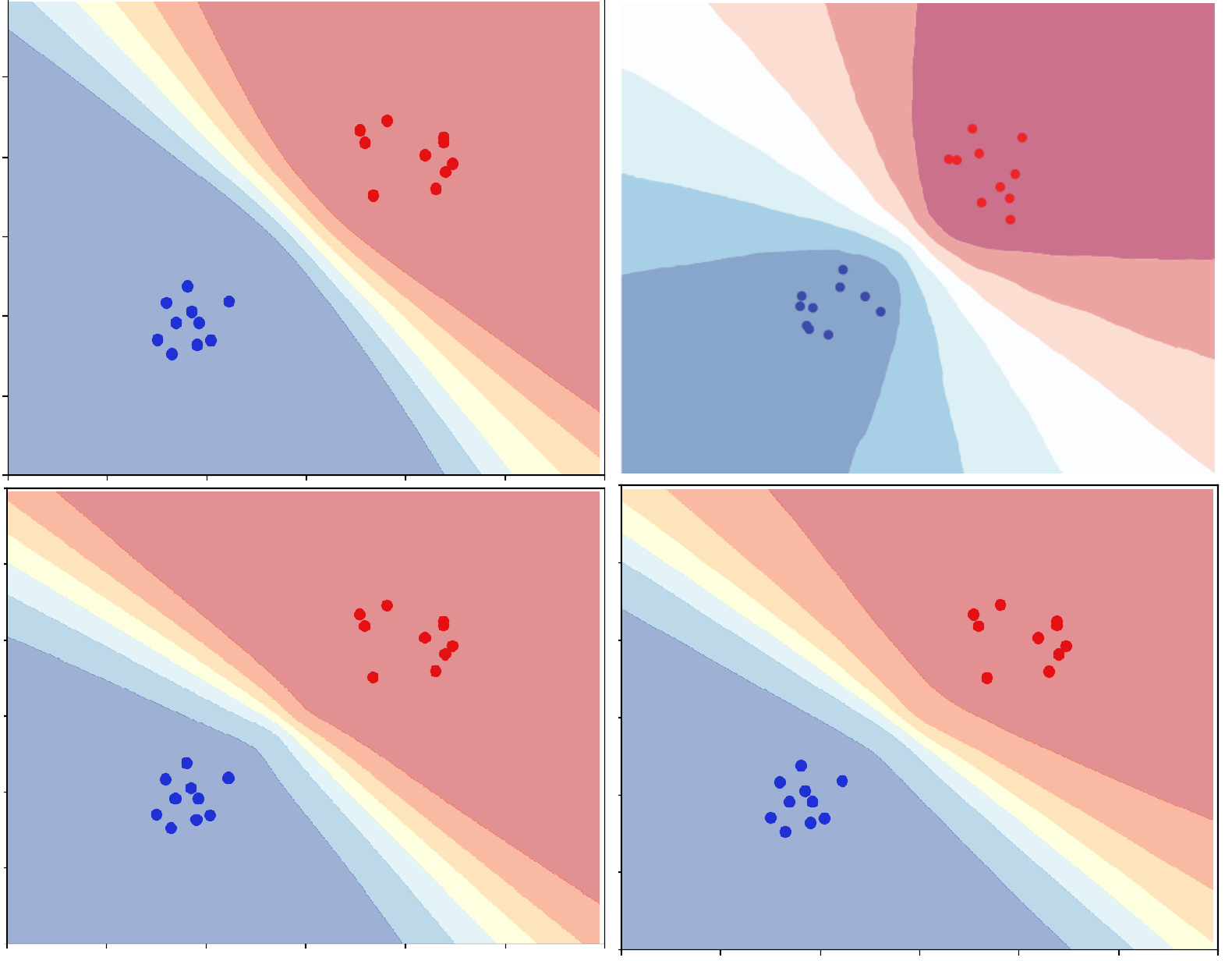}
	\label{fig:toy_results}
	\caption{\small Binary classification on the synthetic dataset, Top left is SVGD, top right is the ground truth, bottom left is PBP\_MV and bottom right is ours. The dots indicate both training and testing data points, different colors illustrates the prediction density, and the higher confidence the model has, the deeper the color will be.}
\end{figure}

We follow the toy example in~\citep{ghosh2016assumed} to consider the binary classification task, in which we uniformly sample 10 data points in seperate distributions:$[-3,-1]\times[-3,-1]$ and $[1,3]\times[1,3]$. A one-layer BNNs with 30 hidden units is employed, we use at most 200 epochs for PBP\_MV, SVGD and S$^2$VGD. The posterior prediction density is plotted in Figure~\ref{fig:toy_results}. The S$^2$VGD performs better than other models, because it is more similar to the ground truth and has a balanced posterior density. PBP\_MV employed the structure information but is a little unbalanced. The SVGD is more unbalanced compared with PBP\_MV.
\begin{figure}[ht]
	\includegraphics[width=1\linewidth]{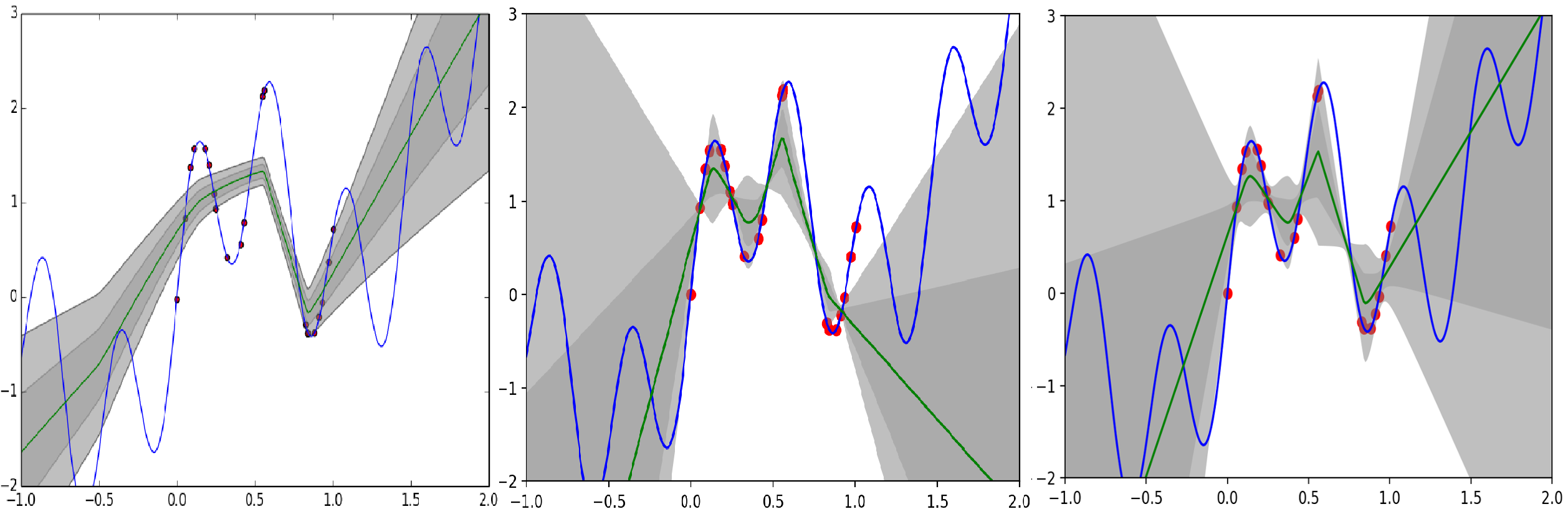}
	\caption{\small Regression on synthetic dataset. From the left to the right are PBP\_MV, SVGD, S$^2$VGD. The blue line is the ground truth. The light grey area shows the $\pm 3$ standard derivation confidence intervals; The green line is the mean of predictions. }
\end{figure}
For non-linear regression, we follow the experiment setup of \citep{louizos2016structured, sun2017learning}. We randomly generated 20 points as input $x_n$, in which 12 points are sampled from Uniform(0,0.6), and 8 points are sampled from Uniform(0.8,1). The output $y_n$ is corresponding to $x_n$, and $y_n = x_n +\epsilon_n + \sin(4(x_n + \epsilon_n)) + sin(13(x_n+\epsilon_n))$, where $\epsilon_n \sim \mathcal{N}(0, 0.0009)$. We fit a one-layer \ReLU~neural network with 100 hidden units. We run PBP\_MV, SVGD and S$^2$VGD for at most 1500 epochs.

Compared with other methods, S$^2$VGD captures uncertainty on two-sides using its variance, but other methods only capture part of the uncertainty.

\vspace{-0mm}
\begin{table*}[ht]\centering \hspace{-0mm}
	\caption{\small Hyper-parameter settings for policy gradient experiments.}
	\vspace{-3mm}
	\label{tab:rl_para}
	\vskip 0.0in
	\centering
	\small
	\hspace{ 0mm} 	
	\begin{adjustbox}{scale=1.0,tabular=l|ccc,center}
		\hline
		{\bf Datasets}
		   &  $\mathtt{Cartpole Swingup}$ &  $\mathtt{Double Pendulum}$ &  $\mathtt{Cartpole}$\\
		\hline
		Batch Size  	  &   5000  &   5000  &   5000  \\
		Step Size &  $5\!\times\!10^{-3}$ &  $5\!\times\!10^{-3}$ &  $5\!\times\!10^{-3}$ \\	
		\#Episodes &1000 & 1000 & 100\\
		Discount  & 0.99 & 0.99 & 0.99\\
		Temperature & [7,8,9,10,11]&[7,8,9,10,11]&[7,8,9,10,11]\\
		\hline
		Network & [25, 10]& [25, 10] & [25, 10] 									  	\\					
		\hline
		Variance in prior & 0.01 & 0.01 & 0.01  \\
		\hline
	\end{adjustbox}
	\vspace{2mm}
\end{table*}

\section{Experimental setting}
We discuss the hyper-parameter settings for S$^2$VGD, and then provide their values for our experiments. All experiments are conducted on a single TITAN X GPU.

\subsection{Discussion of Hyper-parameters}

\paragraph{Step Size} The step size $\epsilon$ for SVGD and S$^2$VGD corresponds to the optimization counterparts. A block decay strategy is used on several datasets, it decreases by the stepsize by half every $L$ epochs.

\paragraph{Mini-batch Size} The gradient at step $t$ is evaluated on a batch of data $\Scal_t$. For small datasets, the batch size can be set to the training sample size $|\Scal_t| = N$, giving the true gradient for each step. For large datasets, a stochastic gradient evaluated from a mini-batch of size  $|\Scal_t| < N$ is used to

\paragraph{Variance of Gaussian Prior} The prior distributions on the parameterized weights of DNNs are Gaussian, with mean 0 and variance $\sigma^2$. The variance of this Gaussian distribution determines the prior belief of how strongly these weights should concentrate on 0. This setting depends on user perception of the amount variability existing in the data. A larger variance in the prior leads to a wider range of weight choices, thus higher uncertainty. The weight decay value of $\ell_2$ regularization in stochastic optimization is related to the prior variance in SVGD and S$^2$VGD.

\paragraph{Number of Particles} The number of particles $M$, is  employed to approximate the posterior.  SVGD and S$^2$VGD represents the posterior approximately in terms of a set of particles (samples), and is endowed with guarantees on the approximation accuracy when the number of particles is exactly infinity~\citep{liu2017stein_flow}.  The number of particles balanced posterior approximation accuracy and computational cost.

\paragraph{Number of Householder Transformations} Instead of maintaining a full orthogonal matrix, we approximate it employing householder flows containing $K$ Householder transformations. According to Lemma~\ref{lem:householder}, the orthogonal matrix can be {\bf exactly} represented by a series of Householder transformations, when $K$ is the degree. See details in Supplementary D. In experiments,  the number of Householder transformations balanced orthogonal matrix approximation accuracy and computational cost.

\subsection{Settings in Our Experiments}
The hyper-parameter settings of SVGD and S$^2$VGD on each dataset is specified in  Table~\ref{tab:mnist_para} for MNIST, Table~\ref{tab:cmab_para} for contextual bandits and Table~\ref{tab:rl_para} for reinforcement learning.

\begin{table}[H]\centering \hspace{-0mm}
	\caption{\small Hyper-parameter settings for contextual bandits.}
	\vspace{-3mm}
	\label{tab:cmab_para}
	\vskip 0.0in
	\centering
	\small
	\hspace{ 0mm} 	
	\begin{adjustbox}{scale=1.0,tabular=l|cc,center}
		\hline
		{\bf Datasets}
		&  $\mathtt{Mushroom}$ &  $\mathtt{Yahoo! Today}$ \\
		\hline
		Batch Size  	  &   64  &   100 \\
		Step Size &  $10^{-3}$ &  $10^{-3}$ \\	
		\#Trial &$5\!\times\!10^{4}$ & $4.5\!\times\!10^{7}$\\
		\hline
		Network (hidden layers) & [50]& [50] 									  	\\					
		\hline
		Variance in prior & 1 & 1   \\
		\hline
	\end{adjustbox}
	\vspace{2mm}
\end{table}

\begin{table}[H]\centering \hspace{-0mm}
	\caption{\small Hyper-parameter settings for MNIST.}
	\vspace{-3mm}
	\label{tab:mnist_para}
	\vskip 0.0in
	\centering
	\small
	\hspace{ 0mm} 	
	\begin{adjustbox}{scale=1.0,tabular=l|cc,center}
		\hline
		{\bf Datasets} &\multicolumn{2}{c}{$\mathtt{MNIST}$}\\
		\hline
		Batch Size  	  &   100  &   100 \\
		Step Size &  $5\!\times10^{-4}$ &  $5\!\times10^{-4}$ \\	
		\#Epoch &$150$ & $150$\\
		RMSProp &0.99 & 0.99\\
		\hline
		Network (hidden layers) & [400, 400]& [800, 800] 	\\					
		\hline
		Variance in prior & 1 & 1   \\
		\hline
	\end{adjustbox}
	\vspace{2mm}
\end{table}
%


%% file: s2vgd.bbl
\begin{thebibliography}{44}
\providecommand{\natexlab}[1]{#1}
\providecommand{\url}[1]{\texttt{#1}}
\expandafter\ifx\csname urlstyle\endcsname\relax
  \providecommand{\doi}[1]{doi: #1}\else
  \providecommand{\doi}{doi: \begingroup \urlstyle{rm}\Url}\fi

\bibitem[Blundell et~al.(2015)Blundell, Cornebise, Kavukcuoglu, and
  Wierstra]{blundell2015weight}
Charles Blundell, Julien Cornebise, Koray Kavukcuoglu, and Daan Wierstra.
\newblock Weight uncertainty in neural networks.
\newblock In \emph{ICML}, 2015.

\bibitem[Chen and Zhang(2017)]{chen2017particle}
Changyou Chen and Ruiyi Zhang.
\newblock Particle optimization in stochastic gradient mcmc.
\newblock \emph{arXiv:1711.10927}, 2017.

\bibitem[Duan et~al.(2016)Duan, Chen, Houthooft, Schulman, and
  Abbeel]{duan2016benchmarking}
Yan Duan, Xi~Chen, Rein Houthooft, John Schulman, and Pieter Abbeel.
\newblock Benchmarking deep reinforcement learning for continuous control.
\newblock In \emph{ICML}, 2016.

\bibitem[Feng et~al.(2018)Feng, Wang, and Liu]{feng2017learning}
Yihao Feng, Dilin Wang, and Qiang Liu.
\newblock Learning to draw samples with amortized stein variational gradient
  descent.
\newblock In \emph{UAI}, 2018.

\bibitem[Fortunato et~al.(2017)Fortunato, Blundell, and
  Vinyals]{fortunato2017bayesian}
Meire Fortunato, Charles Blundell, and Oriol Vinyals.
\newblock Bayesian recurrent neural networks.
\newblock \emph{arXiv:1704.02798}, 2017.

\bibitem[Gal and Ghahramani(2016)]{gal2016dropout}
Yarin Gal and Zoubin Ghahramani.
\newblock Dropout as a bayesian approximation: Representing model uncertainty
  in deep learning.
\newblock In \emph{ICML}, 2016.

\bibitem[Gan et~al.(2017)Gan, Li, Chen, Pu, Su, and Carin]{gan2016scalable}
Zhe Gan, Chunyuan Li, Changyou Chen, Yunchen Pu, Qinliang Su, and Lawrence
  Carin.
\newblock Scalable bayesian learning of recurrent neural networks for language
  modeling.
\newblock 2017.

\bibitem[Ghosh et~al.(2016)Ghosh, Delle~Fave, and Yedidia]{ghosh2016assumed}
Soumya Ghosh, Francesco~Maria Delle~Fave, and Jonathan Yedidia.
\newblock Assumed density filtering methods for learning bayesian neural
  networks.
\newblock In \emph{AAAI}, 2016.

\bibitem[Golub and Van~Loan(2012)]{golub2012matrix}
Gene~H Golub and Charles~F Van~Loan.
\newblock \emph{Matrix Computations}.
\newblock 2012.

\bibitem[Gorham and Mackey(2017)]{gorham2017measuring}
Jackson Gorham and Lester Mackey.
\newblock Measuring sample quality with kernels.
\newblock \emph{arXiv:1703.01717}, 2017.

\bibitem[Gupta and Nagar(1999)]{gupta1999matrix}
Arjun~K Gupta and Daya~K Nagar.
\newblock \emph{Matrix Variate Distributions}.
\newblock 1999.

\bibitem[Hern{\'a}ndez-Lobato and Adams(2015)]{hernandez2015probabilistic}
Jos{\'e}~Miguel Hern{\'a}ndez-Lobato and Ryan Adams.
\newblock Probabilistic backpropagation for scalable learning of bayesian
  neural networks.
\newblock In \emph{ICML}, 2015.

\bibitem[Hinton et~al.(2012)Hinton, Srivastava, and Swersky]{hinton2012rmsprop}
Geoffrey~E Hinton, Nitish Srivastava, and Kevin Swersky.
\newblock Rmsprop: Divide the gradient by a running average of its recent
  magnitude.
\newblock \emph{Neural Networks for Machine Learning, Coursera}, 2012.

\bibitem[Houthooft et~al.(2016)Houthooft, Chen, Duan, Schulman, De~Turck, and
  Abbeel]{houthooft2016vime}
Rein Houthooft, Xi~Chen, Yan Duan, John Schulman, Filip De~Turck, and Pieter
  Abbeel.
\newblock Vime: Variational information maximizing exploration.
\newblock In \emph{NIPS}, 2016.

\bibitem[Kawale et~al.(2015)Kawale, Bui, Kveton, Tran-Thanh, and
  Chawla]{kawale2015efficient}
Jaya Kawale, Hung~H Bui, Branislav Kveton, Long Tran-Thanh, and Sanjay Chawla.
\newblock Efficient thompson sampling for online￼ matrix-factorization
  recommendation.
\newblock In \emph{NIPS}, 2015.

\bibitem[Kingma and Ba(2015)]{kingma2014adam}
Diederik Kingma and Jimmy Ba.
\newblock Adam: A method for stochastic optimization.
\newblock In \emph{ICLR}, 2015.

\bibitem[Kolter and Ng(2009)]{kolter2009near}
J~Zico Kolter and Andrew~Y Ng.
\newblock Near-bayesian exploration in polynomial time.
\newblock In \emph{ICML}, 2009.

\bibitem[Krizhevsky et~al.(2012)Krizhevsky, Sutskever, and
  Hinton]{krizhevsky2012imagenet}
Alex Krizhevsky, Ilya Sutskever, and Geoffrey~E Hinton.
\newblock Imagenet classification with deep convolutional neural networks.
\newblock In \emph{NIPS}, 2012.

\bibitem[Li et~al.(2016{\natexlab{a}})Li, Chen, Carlson, and
  Carin]{li2016preconditioned}
Chunyuan Li, Changyou Chen, David~E Carlson, and Lawrence Carin.
\newblock Preconditioned stochastic gradient langevin dynamics for deep neural
  networks.
\newblock In \emph{AAAI}, 2016{\natexlab{a}}.

\bibitem[Li et~al.(2016{\natexlab{b}})Li, Stevens, Chen, Pu, Gan, and
  Carin]{li2016learning}
Chunyuan Li, Andrew Stevens, Changyou Chen, Yunchen Pu, Zhe Gan, and Lawrence
  Carin.
\newblock Learning weight uncertainty with stochastic gradient mcmc for shape
  classification.
\newblock In \emph{CVPR}, 2016{\natexlab{b}}.

\bibitem[Li et~al.(2010)Li, Chu, Langford, and Schapire]{li2010contextual}
Lihong Li, Wei Chu, John Langford, and Robert~E Schapire.
\newblock A contextual-bandit approach to personalized news article
  recommendation.
\newblock In \emph{WWW}, 2010.

\bibitem[Li et~al.(2011)Li, Chu, Langford, and Wang]{li2011unbiased}
Lihong Li, Wei Chu, John Langford, and Xuanhui Wang.
\newblock Unbiased offline evaluation of contextual-bandit-based news article
  recommendation algorithms.
\newblock In \emph{WSDM}, 2011.

\bibitem[Li et~al.(2015)Li, Hern{\'a}ndez-Lobato, and Turner]{li2015stochastic}
Yingzhen Li, Jos{\'e}~Miguel Hern{\'a}ndez-Lobato, and Richard~E Turner.
\newblock Stochastic expectation propagation.
\newblock In \emph{NIPS}, 2015.

\bibitem[Liu(2017)]{liu2017stein_flow}
Qiang Liu.
\newblock Stein variational gradient descent as gradient flow.
\newblock In \emph{NIPS}, 2017.

\bibitem[Liu and Wang(2016)]{liu2016stein}
Qiang Liu and Dilin Wang.
\newblock Stein variational gradient descent: A general purpose bayesian
  inference algorithm.
\newblock In \emph{NIPS}, 2016.

\bibitem[Liu et~al.(2017)Liu, Ramachandran, Liu, and Peng]{liu2017stein}
Yang Liu, Prajit Ramachandran, Qiang Liu, and Jian Peng.
\newblock Stein variational policy gradient.
\newblock In \emph{UAI}, 2017.

\bibitem[Louizos and Welling(2016)]{louizos2016structured}
Christos Louizos and Max Welling.
\newblock Structured and efficient variational deep learning with matrix
  gaussian posteriors.
\newblock In \emph{NIPS}, 2016.

\bibitem[MacKay(1992)]{mackay1992practical}
David~JC MacKay.
\newblock A practical bayesian framework for backpropagation networks.
\newblock \emph{Neural Computation}, 1992.

\bibitem[Nair and Hinton(2010)]{nair2010rectified}
Vinod Nair and Geoffrey~E Hinton.
\newblock Rectified linear units improve restricted boltzmann machines.
\newblock In \emph{ICML}, 2010.

\bibitem[Oates et~al.(2016)Oates, Cockayne, Briol, and
  Girolami]{oates2016convergence}
Chris~J Oates, Jon Cockayne, Fran{\c{c}}ois-Xavier Briol, and Mark Girolami.
\newblock Convergence rates for a class of estimators based on stein's
  identity.
\newblock \emph{arXiv:1603.03220}, 2016.

\bibitem[Pu et~al.(2017{\natexlab{a}})Pu, Chen, Dai, Wang, Li, and
  Carin]{pu2017symmetric}
Yunchen Pu, Liqun Chen, Shuyang Dai, Weiyao Wang, Chunyuan Li, and Lawrence
  Carin.
\newblock Symmetric variational autoencoder and connections to adversarial
  learning.
\newblock 2017{\natexlab{a}}.

\bibitem[Pu et~al.(2017{\natexlab{b}})Pu, Gan, Henao, Li, Han, and
  Carin]{pu2017stein}
Yunchen Pu, Zhe Gan, Ricardo Henao, Chunyuan Li, Shaobo Han, and Lawrence
  Carin.
\newblock Stein variational autoencoder.
\newblock In \emph{NIPS}, 2017{\natexlab{b}}.

\bibitem[Russo et~al.(2017)Russo, Tse, and Van~Roy]{russo2017time}
Daniel Russo, David Tse, and Benjamin Van~Roy.
\newblock Time-sensitive bandit learning and satisficing thompson sampling.
\newblock \emph{arXiv:1704.09028}, 2017.

\bibitem[Schulman et~al.(2016)Schulman, Moritz, Levine, Jordan, and
  Abbeel]{schulman2015high}
John Schulman, Philipp Moritz, Sergey Levine, Michael Jordan, and Pieter
  Abbeel.
\newblock High-dimensional continuous control using generalized advantage
  estimation.
\newblock In \emph{ICLR}, 2016.

\bibitem[Silver et~al.(2016)Silver, Huang, et~al.]{silver2016mastering}
David Silver, Aja Huang, et~al.
\newblock Mastering the game of go with deep neural networks and tree search.
\newblock \emph{Nature}, 2016.

\bibitem[Sun et~al.(2017)Sun, Chen, and Carin]{sun2017learning}
Shengyang Sun, Changyou Chen, and Lawrence Carin.
\newblock Learning structured weight uncertainty in bayesian neural networks.
\newblock In \emph{AISTATS}, 2017.

\bibitem[Sun and Bischof(1995)]{sun1995basis}
Xiaobai Sun and Christian Bischof.
\newblock A basis-kernel representation of orthogonal matrices.
\newblock \emph{SIAM Journal on Matrix Analysis and Applications}, 1995.

\bibitem[Sutskever et~al.(2014)Sutskever, Vinyals, and
  Le]{sutskever2014sequence}
Ilya Sutskever, Oriol Vinyals, and Quoc~V Le.
\newblock Sequence to sequence learning with neural networks.
\newblock In \emph{NIPS}, 2014.

\bibitem[Sutton and Barto(1998)]{sutton1998reinforcement}
Richard~S Sutton and Andrew~G Barto.
\newblock \emph{Reinforcement learning: An introduction}.
\newblock 1998.

\bibitem[Thompson(1933)]{thompson1933likelihood}
William~R Thompson.
\newblock On the likelihood that one unknown probability exceeds another in
  view of the evidence of two samples.
\newblock \emph{Biometrika}, 1933.

\bibitem[Tomczak and Welling(2016)]{tomczak2016improving}
Jakub~M Tomczak and Max Welling.
\newblock Improving variational auto-encoders using householder flow.
\newblock \emph{arXiv:1611.09630}, 2016.

\bibitem[Williams(1992)]{williams1992simple}
Ronald~J Williams.
\newblock Simple statistical gradient-following algorithms for connectionist
  reinforcement learning.
\newblock \emph{Machine Learning}, 1992.

\bibitem[Zhang et~al.(2016)Zhang, Wang, Chen, Henao, Fan, and
  Carin]{zhang2016towards}
Yizhe Zhang, Xiangyu Wang, Changyou Chen, Ricardo Henao, Kai Fan, and Lawrence
  Carin.
\newblock Towards unifying hamiltonian monte carlo and slice sampling.
\newblock In \emph{NIPS}, 2016.

\bibitem[Zhang et~al.(2017)Zhang, Chen, Gan, Henao, and
  Carin]{zhang2017stochastic}
Yizhe Zhang, Changyou Chen, Zhe Gan, Ricardo Henao, and Lawrence Carin.
\newblock Stochastic gradient monomial gamma sampler.
\newblock In \emph{ICML}, 2017.

\end{thebibliography}
